\documentclass[10pt,twocolumn,letterpaper]{article}

\usepackage{3dv}
\usepackage{times}
\usepackage{epsfig}
\usepackage{graphicx}
\usepackage{amsmath}
\usepackage{amssymb}
\usepackage{enumitem} 
\usepackage{euscript} 
\usepackage{booktabs} 
\usepackage{diagbox} 

\usepackage[pagebackref=true,breaklinks=true,letterpaper=true,colorlinks,bookmarks=false]{hyperref}

\threedvfinalcopy 

\def\threedvPaperID{243} 

\ifthreedvfinal\fi

\long\def\ignorethis#1{}

\usepackage{color}
\definecolor{gray}{rgb}{0.35,0.35,0.35}
\definecolor{red}{rgb}{1,0,0}
\definecolor{dark-green}{rgb}{0,0.4,0}
\definecolor{blue}{rgb}{0,0,1}
\definecolor{orange}{rgb}{1,0.55,0}
\definecolor{white}{rgb}{1,1,1}
\definecolor{black}{rgb}{0,0,0}
\definecolor{dark-brown}{rgb}{0.2,0.1,0}
\definecolor{light-blue}{rgb}{0.4,0.6,0.99}
\definecolor{dark-red}{rgb}{0.6,0,0}
\definecolor{light-red}{rgb}{1,0.2,0.6}
\definecolor{pink}{rgb}{1,0.2,0.6}
\definecolor{dark-pink}{rgb}{0.6,0,0.3}

\newcommand{\whitetxt}[1]{{\color{white}#1}\normalfont}
\newcommand{\newtxt}[1]{{\color{black}#1}\normalfont}

\newbox\jsavebox

\newcommand{\Lagr}{\mathcal{L}}

\mathchardef\mhyphen="2D
\newcommand{\SCD}{S \mhyphen CD}
\newcommand{\SRE}{S \mhyphen RE}
\newcommand{\SRCA}{S \mhyphen RCA}
\newcommand{\SNRE}{S \mhyphen NRE}
\newcommand{\TRE}{T \mhyphen RE}
\newcommand{\TNRE}{T \mhyphen NRE}

\DeclareMathOperator*{\argmin}{\arg\!\min}

\begin{document}

\title{Geometric Adversarial Attacks and Defenses on 3D Point Clouds}

\author{Itai Lang\\
Tel Aviv University\\
{\tt\small itailang@mail.tau.ac.il}
\and
Uriel Kotlicki\\
Tel Aviv University\\
{\tt\small kotlicki@mail.tau.ac.il}
\and
Shai Avidan\\
Tel Aviv University\\
{\tt\small avidan@eng.tau.ac.il}
}

\maketitle
\thispagestyle{empty} 

\begin{abstract}
Deep neural networks are prone to adversarial examples that maliciously alter the network's outcome. Due to the increasing popularity of 3D sensors in safety-critical systems and the vast deployment of deep learning models for 3D point sets, there is a growing interest in adversarial attacks and defenses for such models. So far, the research has focused on the semantic level, namely, deep point cloud classifiers. However, point clouds are also widely used in a geometric-related form that includes encoding and reconstructing the geometry.

In this work, we are the first to consider the problem of adversarial examples at a geometric level. In this setting, the question is how to craft a small change to a clean source point cloud that leads, after passing through an autoencoder model, to the reconstruction of a different target shape. Our attack is in sharp contrast to existing semantic attacks on 3D point clouds. While such works aim to modify the predicted label by a classifier, we alter the entire reconstructed geometry. Additionally, we demonstrate the robustness of our attack in the case of defense, where we show that remnant characteristics of the target shape are still present at the output after applying the defense to the adversarial input. Our code is publicly available\footnote{\url{https://github.com/itailang/geometric_adv}}.


\end{abstract}




\section{Introduction} \label{sec:introduction}
A point cloud is an important 3D data representation. It is lightweight in memory, simple in form, and very common as the output format of a 3D sensor. In recent years, deep learning methods have been very successful in processing point clouds. Applications range from high-level tasks, such as classification~\cite{qi2017pointnet, qi2017pointnetpp, wang2019dynamic}, semantic segmentation~\cite{landrieu2018large, tchapmi2017segcloud, wang2019graph}, and object detection~\cite{qi2019deep, shi2019pointrcnn, yang20203dssd}, to low-level tasks, such as down- and up-sampling~\cite{dovrat2019learning, lang2020samplenet, li2019pu-gan, yu2018pu-net}, denoising~\cite{hermosilla2019total, rakotosaona2020pointcleannet}, and reconstruction~\cite{achlioptas2018learning, groueix2018atlasnet, yang2018folding}.

Despite their great success, neural networks are vulnerable to adversarial attacks. This topic has drawn much attention in the case of 2D images, where the main focus has been on the \textit{semantic} level, namely, adversarial examples that mislead image classifiers~\cite{carlini2017towards, goodfellow2015explaining, papernot2016the}.

\begin{figure}[t!]
\begin{center}
\includegraphics[width=0.95\columnwidth]{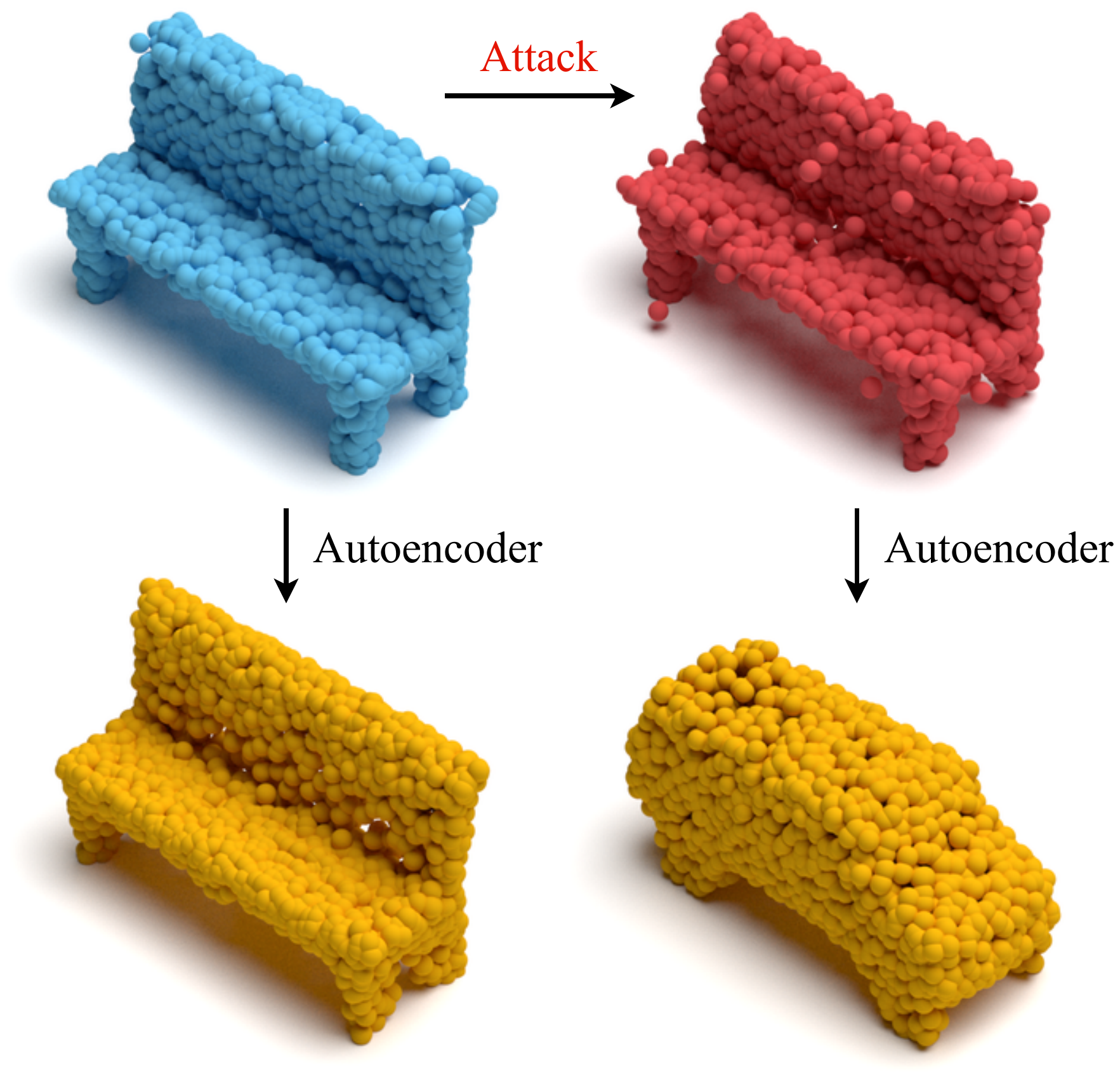}
\caption{{\bfseries A geometric adversarial attack.} Given a clean input point cloud (top left), our attack crafts an adversarial example (top right) that changes the \textit{geometry} at the output of an \textit{autoencoder} model (bottom row). In this case, it turns a \textit{bench} into a \textit{car}.}
\label{fig:teaser}
\end{center}
\end{figure}




Very recently, semantic adversarial attacks have been extended to 3D point cloud classification~\cite{xiang2019generating, zheng2019pointcloud, zhou2020lg-gan} and 3D object detection in autonomous vehicles~\cite{cao2019adversarial, sun2020towards, tu2020physically}. Still, safety-critical applications that process 3D data, including object manipulation by a robotic arm, obstacle avoidance, and road navigation, should have a precise \textit{geometric} perception of its environment for a desirable safe operation.

Since a point cloud is a geometric entity, a natural question is whether an adversarial attack is possible at the \textit{geometric} level. Meaning, can we make a small change to an input point cloud, feed it through an autoencoder network, and reconstruct a \textit{geometrically different} output? This problem has not been studied before.

This paper suggests a framework for \textit{geometric} adversarial attacks on 3D point clouds. We assume that a point cloud is to be processed by an autoencoder (AE) model. The encoder and decoder can be separate, for example, at a client unit and a server, respectively; alternatively, both parts of the AE may be on the same device, and the AE compresses the data for storage or communication.



\newtxt{Given a clean source point cloud, our attack changes the reconstructed geometry by the AE to a different target shape}. Two attack variants are explored: one is a gray-box attack that only has access to the encoder and operates on the latent space of the AE; the other is a white-box attack that has access to both the encoder and decoder parts and operates on the output space. The first attack crafts an adversarial example that is encoded to a target latent vector. The second one minimizes the reconstruction error with respect to a desired output shape (see an example in Figure~\ref{fig:teaser}). In addition, we evaluate the transferability of our attacks to other AE models. 

We also examine \newtxt{the attack's robustness to} two types of defenses. One defense removes off-surface points, which are an artifact of our attack. These points are shifted by the attack out of the shape's surface and seem to be floating around it. This defense is architecture-independent. The other defense drops points that are critical to the malicious reconstruction. It depends on the architecture of the AE that we attack~\cite{achlioptas2018learning}. \newtxt{We use the source distortion and target reconstruction error as geometric metrics to measure the success of our attacks.} Additionally, we evaluate the semantic interpretation of reconstructed adversarial and defended point sets.


Our experiments demonstrate the effectiveness of geometric attacks in the sense that a small perturbation to a clean point set causes the AE to output a different shape with a low reconstruction error. We also show that the attacks are not entirely defendable. Some features of the target shape survive the defense and appear in the reconstruction of the defended point set.





To summarize, our main contributions in this paper are as follows:
\newtxt{
\begin{itemize}[noitemsep, nolistsep]
    \item We explore the problem of geometric adversarial attacks on 3D point clouds. It is in contrast to existing attacks on point cloud classifiers, targeting to alter the perceived semantic meaning of the object. We, on the other hand, completely change the reconstructed geometry by a victim AE;
    \item We extensively evaluate the adversarial impact of our proposed attacks and their resistance to counter defenses, demonstrating the effectiveness and robustness of the attacks.
\end{itemize}
}

\section{Related Work} \label{sec:related_work}
\noindent {\bfseries Representation learning for point sets} \quad Nowadays, representation learning for 3D data is a prevalent research topic~\cite{li2018so-net, mescheder2019occupancy, mo2019structurenet, zhao20193d}. A ubiquitous approach is to employ an encoder-decoder network architecture, typically referred to as an autoencoder (AE)~\cite{hinton2006reducing}. The encoder encodes the essence of the input to a latent code-word. The decoder learns a transformation from the latent space back to the raw space to reconstruct the input. Another typical flavor is a variational autoencoder (VAE)~\cite{kingma2014auto}, in which a statistical constraint is imposed on the learned latent space. In addition to reconstruction, an AE or VAE can be used for other applications, such as shape classification~\cite{yang2018folding}, morphing~\cite{groueix2018atlasnet}, editing~\cite{mehr2019disconet, mo2020structedit}, analogies~\cite{achlioptas2018learning}, and synthesis~\cite{chen2019learning, gao2019sdm-net}, in an intuitive and elegant manner.


A variety of AEs have been proposed for point clouds. Achlioptas \etal~\cite{achlioptas2018learning} proposed an AE that operates directly on the 3D coordinates of the input. Its architecture, based on PointNet's architecture~\cite{qi2017pointnet}, employs a per-point multi-layer perception (MLP) and a global max-pooling operation to obtain the latent representation of the input set. Then, several fully-connected layers regress the $xyz$ coordinates back. A follow-up work enriched the encoder by aggregating information from neighboring points~\cite{yang2018folding}. Researchers also suggested alternatives for the decoder, such as folding one or multiple 2D patches onto the 3D surface of the shape~\cite{groueix2018atlasnet, yang2018folding}. In our work, we attack the widespread AE of Achlioptas~\etal~\cite{achlioptas2018learning} and check the attack transferability to other common AEs, AtlasNet~\cite{groueix2018atlasnet}, and FoldingNet~\cite{yang2018folding}.

\medskip
\noindent {\bfseries Adversarial attacks} \quad An adversarial attack creates an example that compromises a victim network's behavior at the attacker's will. Crafting an adversarial example is typically cast as an optimization problem, with an adversarial loss on the network's output, along with a regularization term on the distortion of the input. Adversarial attacks have been thoroughly studied for 2D image classifiers~\cite{carlini2017towards, goodfellow2015explaining, papernot2016the, su2019one} and lately have been expanded to 3D point clouds~\cite{hamdi2020advpc, wen2020geometry, xiang2019generating, zhao2020on}.



Xiang \etal~\cite{xiang2019generating} pioneered adversarial attacks on the PointNet~\cite{qi2017pointnet} classification model. They proposed two types of attacks: one that perturbs existing points, and the other that adds points to a clean example in a cluster or a pre-defined shape. Zheng \etal~\cite{zheng2019pointcloud} and Zhang \etal~\cite{zhang2019adversarial} dropped points that influence the classifier's label prediction in order to change it. Lee \etal~\cite{lee2020shapeadv} injected adversarial noise to the latent space of an AE instead of performing the attack at the point space. The decoded shape resembled the original one while serving as an adversarial example to a subsequent recognition model. Previous works have suggested to attack classification networks for point sets to change their prediction. In contrast, we target an AE model and aim to alter the reconstructed shape.




\medskip
\noindent {\bfseries Defense methods} \quad Defense against adversarial attacks tries to mitigate the attack's harmful effect. Several defense methods have been proposed for protecting 3D point cloud classifiers~\cite{dong2020self, zhang2019adversarial, zhou2019dup-net}. DUP-Net~\cite{zhou2019dup-net} removed out-of-surface points and increased the point cloud's resolution to enable its correct classification. Liu \etal~\cite{liu2019extending} used adversarial examples during training for improving the robustness against such examples. Yang \etal~\cite{zhang2019adversarial} measured categorization stability under random perturbations to detect offensive inputs. Unlike our work, defense methods for point sets were studies for semantic point cloud recognition, not geometric reconstruction, as suggested in this paper.


In our work, we adopt an off-surface point filtering approach as a defense~\cite{liu2019extending, zhou2019dup-net}. Additionally, since the attacked AE~\cite{achlioptas2018learning} employs the PointNet architecture, it applies the critical points theorem~\cite{qi2017pointnet}. Meaning, a subset of the input point cloud, denoted as critical points, determines its reconstructed output. While this notion has been used in the past for attacks~\cite{xiang2019generating, zhang2019adversarial, zheng2019pointcloud}, we leverage it here for defense. 


\section{Method}  \label{sec:method}
We attack an autoencoder (AE) trained on a collection of shapes from several semantic shape classes. We assume that a clean source point cloud is given for the attack. Our goal is to reconstruct a \textit{shape instance} from a different target shape class, with minimal distortion to the source \textit{and} minimal reconstruction error of the target. The target class has various instances with the same semantic type, and we have the freedom to choose from them.

A na\"ive approach is to select the target instance at random. However, this approach is likely to fail since the AE is sensitive to the geometry at its input. \newtxt{Instead, we introduce a \textit{key geometric consideration} in our attack and choose a target shape that is geometrically similar to the source.} Thus, for the given source, we select from the target shape class its nearest neighbor point clouds, in the sense of Chamfer Distance.



If only the encoder is known to the attacker, we treat it as a gray-box attack and work with the latent space of the AE. In a white-box setting, where both the encoder and decoder are accessible, we propose an output space attack that leverages the reconstructed point cloud by the AE.

\subsection{Attacks} \label{subsec:attacks}
\begin{figure}[tb!]
\begin{center}
\includegraphics[width=\columnwidth]{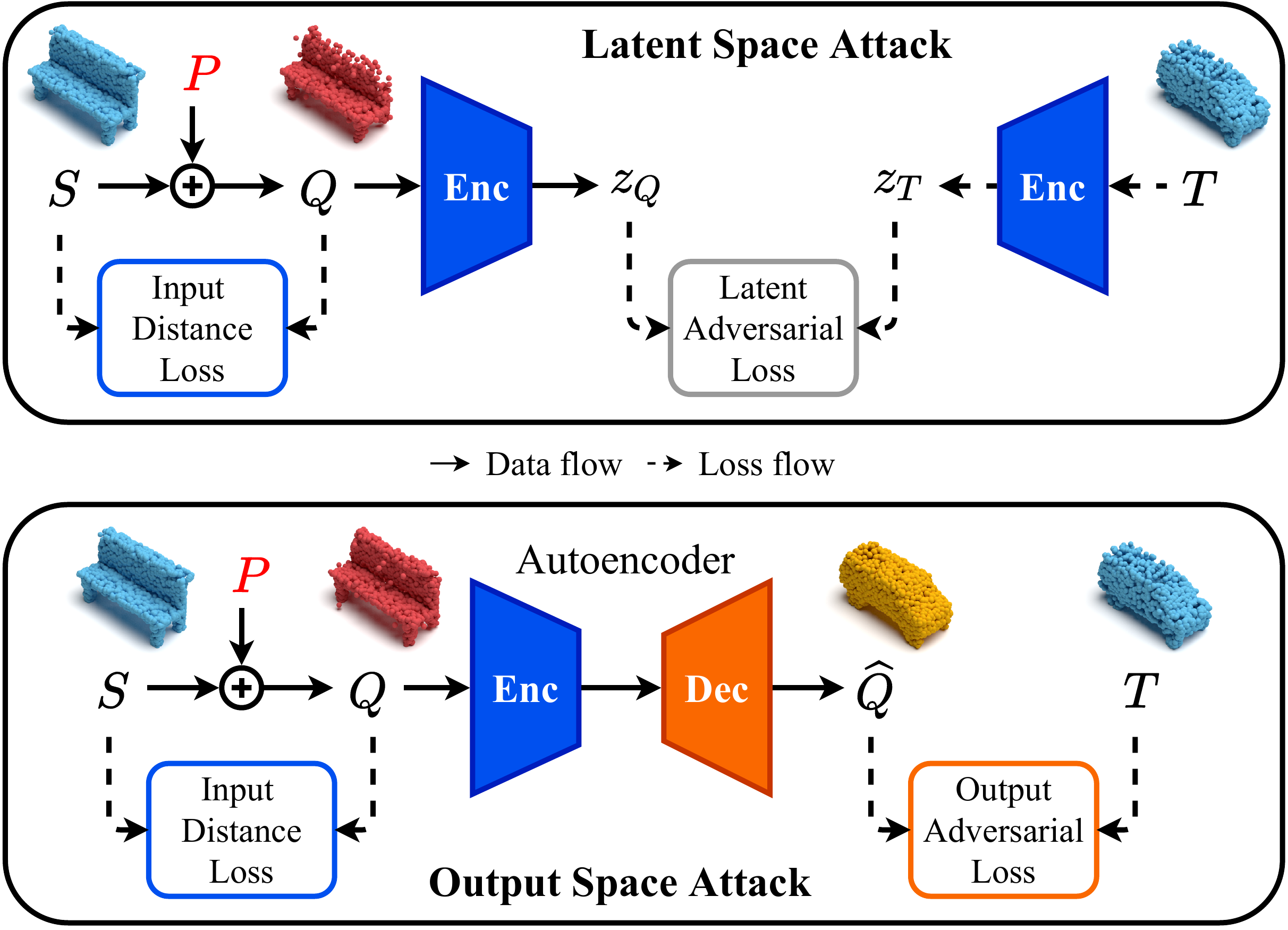}
\caption{{\bfseries The proposed attacks.} An adversarial perturbation $P$ is added to a clean source point cloud $S$, resulting in a malicious input $Q$. In the latent space attack, $P$ is optimized such that $Q$ is encoded to a target latent vector $z_T$ (top), while the output space attack aims to reconstruct a target point cloud $T$ directly (bottom). In both cases, $Q$ is required to resemble the source $S$.}
\label{fig:diagram}
\end{center}
\end{figure}

A diagram of the proposed attacks is presented in Figure~\ref{fig:diagram}. A point cloud is defined as an unordered set of 3D coordinates $S \in \mathbb{R}^{n \times 3}$, where $n$ is the number of points. Given a clean source shape $S$, we add an adversarial perturbation $P$ to obtain an adversarial input $Q$:
\begin{equation} \label{eq:adv_input}
Q = S + P.
\end{equation}

\medskip
\noindent {\bfseries Latent space attack} \quad We cast the attack as the following optimization problem. Given: $S \in \EuScript{S}$, a point cloud from a source shape class; $\EuScript{T}$, a semantically different target shape class; $f_{Enc}$, an encoder model. Find: $Q^* = S + P^*$, an adversarial example that is similar to $S$, and its encoding by $f_{Enc}$ is close to a latent code of a point cloud instance $T$ from $\EuScript{T}$:
\begin{equation} \label{eq:problem_def_latent}
\begin{gathered}
P^* = \argmin_P \; \Lagr_{latent}(z_Q, z_T) + \lambda \Lagr_{distance}(Q, S) \\
s.t. \quad z_Q = f_{Enc}(Q), \quad z_T = f_{Enc}(T), \quad T \in \EuScript{T},
\end{gathered}
\end{equation}

\noindent where $S, \: Q, \: T \in \mathbb{R}^{n \times 3}$, $z_Q, \: z_T \in \mathbb{R}^{m}$, $\EuScript{T} \neq \EuScript{S}$, and $\lambda$ is a hyperparameter. 

This attack's motivation is that proximity in the latent space can result in proximity in the output point space on the decoder side~\cite{yang2019pointflow, yang2018folding}. Due to the Euclidean properties of the latent space~\cite{achlioptas2018learning}, we use Euclidean loss between the latent vectors:
\begin{equation} \label{eq:loss_latent}
\Lagr_{latent}(z_Q,z_T) = ||z_Q - z_T||_2.
\end{equation}


To regularize the attack and maintain similarity between $Q$ and $S$, we use Chamfer Distance, a popular similarity measure between point clouds~\cite{achlioptas2018learning, haoqiang2017apointset, wang2020cascaded, xiang2019generating}. The Chamfer Distance between two point clouds $X$ and $Y$ is given by:
\begin{equation} \label{eq:chamfer_dist}
\begin{split}
C&D(X, Y) = \\
  &\frac{1}{|X|}\sum_{x \in X}{\min_{y \in Y}||x - y||_2^2} +
   \frac{1}{|Y|}\sum_{y \in Y}{\min_{x \in X}||y - x||_2^2}.
\end{split}
\end{equation}

\noindent Thus, the loss for the source and adversarial point clouds is:
\begin{equation} \label{eq:loss_dist}
\Lagr_{distance}(Q, S) = CD(Q, S).
\end{equation}

\medskip
\noindent {\bfseries Output space attack} \quad In case the adversary has access to the entire AE model, we suggest to optimize the model's output. For that attack, the problem statement in Equation~\ref{eq:problem_def_latent} is changed as follows:
\begin{equation} \label{eq:problem_def_output}
\begin{gathered}
P^* = \argmin_P \; \Lagr_{output}(\widehat{Q}, T) + \lambda \Lagr_{distance}(Q, S) \\
s.t. \quad \widehat{Q} = f_{AE}(Q), \quad T \in \EuScript{T},
\end{gathered}
\end{equation}

\noindent where $f_{AE}$ is the AE model, $\widehat{Q}$ is the reconstruction of $Q$ by $f_{AE}$, $\EuScript{T} \neq \EuScript{S}$, and $\lambda$ is a hyperparameter.

The distance loss $\Lagr_{distance}$ is the same as in Equation~\ref{eq:loss_dist}. To obtain the desired target shape $T$, we use Chamfer Distance as the adversarial loss:
\begin{equation} \label{eq:loss_output}
\Lagr_{output}(\widehat{Q},T) = CD(\widehat{Q},T).
\end{equation}

\medskip
\noindent {\bfseries Evaluation metrics} \quad Our attacks' focus is to change the geometry of the shape at the output of the AE network. Thus, we use geometric measures to evaluate the performance of the attacks. For the output, we report the \textit{target} reconstruction error $\TRE = CD(\widehat{Q},T)$. Since the AE model has an inherent error, we are also interested in the relative attack performance. Thus, we use the target normalized reconstruction error~\cite{dovrat2019learning, lang2020samplenet}, defined as:
\begin{equation} \label{eq:t_nre}
\TNRE = \frac{CD(\widehat{Q},T)}{CD(\widehat{T},T)},
\end{equation}

\noindent where $\widehat{T} = f_{AE}(T)$.

For the input, we measure the distortion of the source shape by $\SCD = CD(Q,S)$. In addition, we observed that one artifact of our attacks is off-surface points, denoted as $OS$, in the adversarial point cloud $Q$ (see Figure~\ref{fig:attack} for a visualization). These points are noticeable to a human observer, and thereby, we wish their number to be minimal. We define $OS$ points as points in $Q$ whose distance to the nearest neighbor in $S$ is larger than a threshold:
\begin{equation} \label{eq:off_surface}
OS = \{q \in Q \;|\; \min_{s \in S}||q - s||_2 > \gamma\}.
\end{equation}

\noindent Our goal is to optimize the attacks to achieve \textit{both} minimal input and output metrics.

\newtxt{As a \textit{side-effect}, our geometric attack also induces a semantic attack on the decoder side.} Thus, in addition to the geometric measures, we evaluate the semantic interpretation of the reconstructed point clouds. Meaning, we check whether the adversarial examples' reconstruction can mislead a classifier to the target shape class or avoid the source's class label.


\medskip
\noindent {\bfseries Scoring an attack} \quad The attack must reconcile two contradictory demands. On the one hand, it requires the perturbation of the source shape to be small. On the other, the reconstruction of the adversarial example has to match a true different target shape. 


To cope with this challenge, we select candidate target instances that are close geometrically to the source, as explained at the beginning of this section. We define a score:
\begin{equation} \label{eq:attack_score}
r = CD(Q,S) + CD(\widehat{Q},T),
\end{equation}

\noindent and choose for each source the target instance that yields a minimal score. This process assumes access to the source distortion and target reconstruction error and improves the attack's success in both of them.



\subsection{Defenses} \label{subsec:defenses}
The defense methods operate on the encoder side, on the malicious input point cloud. They intend to deflect adversarial points to reduce the attack's effect on the reconstructed output. \newtxt{We use the defenses to evaluate the robustness of our attacks in such a case.} Two types of defenses are examined. One relies on nearest neighbor statistics in the point cloud to remove off-surface points. This defense does not depend on the attacked AE model. The second defense takes advantage of the victim AE's architecture. It identifies critical points that cause the adversarial reconstruction and filters them out.


\medskip
\noindent {\bfseries Off-surface point removal} \quad In a clean point set, a point typically has neighboring points in its vicinity, on the shape's surface. In contrast, an adversarial input may have points out-of-surface, which change the AE's output.


To detect off-surface points, we compute the average distance to the $k$ nearest neighbors for each point $q \in Q$:
\begin{equation} \label{eq:knn_dist}
\bar{d}_q = \frac{1}{k} \sum_{i \in \EuScript{N}_k(q)}{||q - q_i||_2},
\end{equation}


\noindent where $\EuScript{N}_k(q)$ includes the indices of the $k$ closest points to $q$ in $Q$, excluding $q$. Then, we keep points with an average distance less than or equal to a threshold $\delta$ and obtain the defended point cloud:
\begin{equation} \label{eq:surf_def}
Q^{def}_{surf} = \{q \in Q \;|\; \bar{d}_q \leq \delta\}.
\end{equation}

\noindent Thus, points that do not have close neighbors are removed.

\medskip
\noindent {\bfseries Critical points removal} \quad The adversarial perturbation shifts points to locations that cause the AE to output the desired target shape. We leverage the architecture of the victim AE, which is based on PointNet~\cite{qi2017pointnet}, and identify these points as the critical points \cite{qi2017pointnet} that determine the latent vector of the AE. The critical points defense takes the complementary points, thus reducing the effect of the attack on the reconstructed set. We denote the resulting point cloud of this defense as $Q^{def}_{comp}$. 

\medskip
\noindent {\bfseries Evaluation metrics} \quad We evaluate defense related metrics before and after applying the defense to the adversarial point set. Since the defense's goal is to mitigate the attack's effect, we measure reconstruction errors with respect to the \textit{source} point set. The errors are given by $\SRE_{before} = CD(\widehat{Q},S)$ and $\SRE_{after} = CD(\widehat{Q}^{def},S)$, where $\widehat{Q}^{def}$ is either $f_{AE}(Q^{def}_{surf})$ or $f_{AE}(Q^{def}_{comp})$.

Similar to the attack evaluation (Equation~\ref{eq:t_nre}), we factor in the AE's error and also report the source normalized reconstruction error:
\begin{equation} \label{eq:s_nre_before}
\SNRE_{before} = \frac{CD(\widehat{Q},S)}{CD(\widehat{S},S)}
\end{equation}

\begin{equation} \label{eq:s_nre_after}
\SNRE_{after} = \frac{CD(\widehat{Q}^{def},S)}{CD(\widehat{S},S)},
\end{equation}

\noindent where $\widehat{S} = f_{AE}(S)$.

\section{Results} \label{sec:resutls}
\subsection{Experimental Setup} \label{subsec:experimental_setup}
We evaluate our attacks on point sets of $n = 2048$ points, sampled from ShapeNet Core55 database~\cite{chang2015shapenet}. This database is commonly used for 3D AE research~\cite{achlioptas2018learning}. The point sets are normalized to the unit cube. We set the $OS$ threshold $\gamma = 0.05$, \ie, 5\% of the cube's size. We use the $13$ largest shape classes from the database, with more than 1000 examples each. Every class is split into 85\%/5\%/10\% for train/validation/test sets. $\lambda$ is set to $150$ and $1$ for the latent and output attacks, respectively. \newtxt{We also apply our attack to another dataset~\cite{wu2015modelnet}, as detailed in the supplementary.}


We attack the prevailing point cloud AE of Achlioptas \etal~\cite{achlioptas2018learning}. The model is trained with the authors' recommended settings, and then its parameters are frozen. The source and target shapes for our attacks are all taken from the test set. We select 25 point clouds at random from each shape class as sources for the attack. Each source attacks the other 12 shape classes. For each source, we take 5 point sets as candidates for the attack from the target class, run the attack, and select one target instance per source, as explained at the end of sub-section~\ref{subsec:attacks}. Thus, in total, we obtain $25 \cdot 13 \cdot 12 = 3900$ source-target pairs.

The attacks are implemented on an Nvidia Titan Xp GPU. We optimize them with an Adam optimizer, learning rate 0.01, momentum 0.9, and 500 gradient steps. Additional optimization parameters appear in the supplementary.

\begin{figure*}[tb!]
\begin{center}
\includegraphics[width=0.8\linewidth]{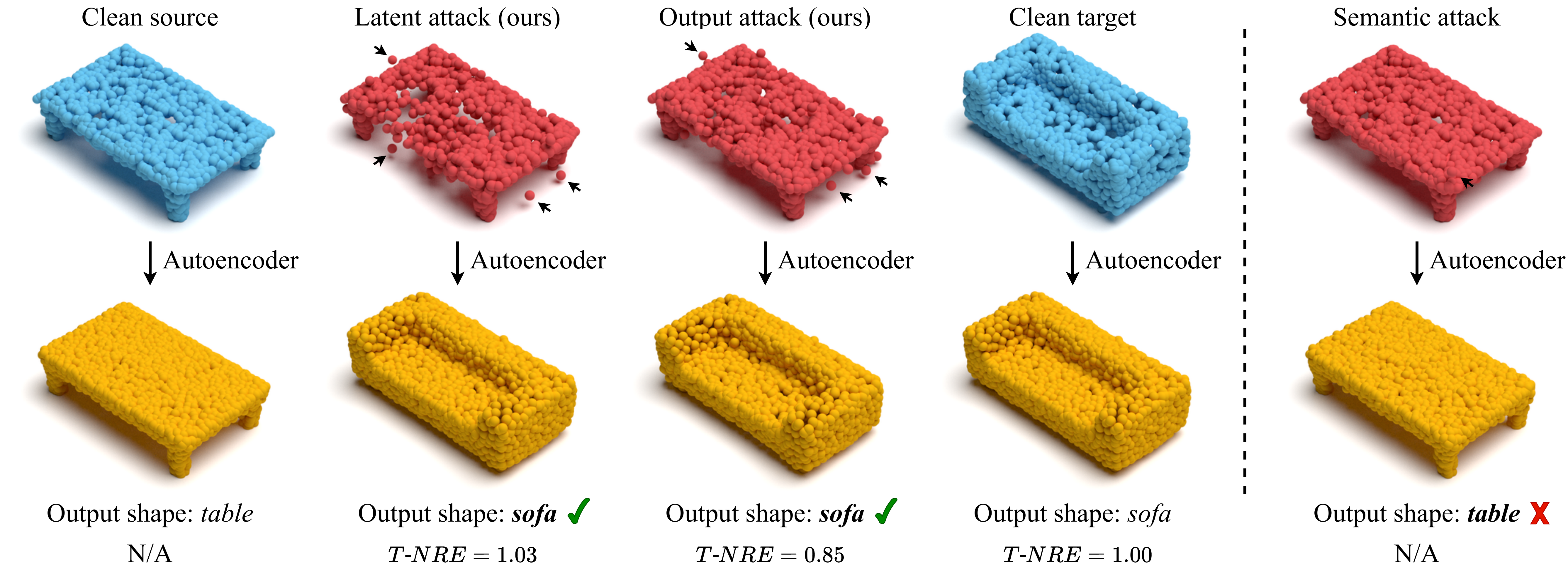}
\caption{{\bfseries Attacks comparison.} Left: geometric attacks. Top row: a clean source point cloud (\textit{table}), adversarial examples of the latent and output space attacks, and a clean target (\textit{sofa}). Bottom row: corresponding reconstructions. Below them, we indicate the class of the attack's target point cloud, the semantic interpretation of the AE's output, and the target normalized reconstruction error, $\TNRE$ (Equation~\ref{eq:t_nre}). We mark off-surface ($OS$) points for the attacks with arrows (part of them, to avoid clutter). Both our attacks alter the reconstructed geometry to the desired target shape instance, while the output attack results in a smaller perturbation of the input point set compared to the latent attack. Right: a semantic adversarial example~\cite{xiang2019generating} and its reconstruction by the AE. In contrast to our attacks, the semantic attack, which misleads a classifier to the \textit{sofa} label, fails to attack the AE and is \textit{not} reconstructed as a sofa but rather as a table.}
\label{fig:attack}
\end{center} 
\end{figure*}

\subsection{Evaluation in Adversarial Setting} \label{subsec:attack_evaluation}
We evaluate the attacks in three aspects: geometric performance, semantic interpretation, and transferability. The first includes the adversarial metrics, as explained in sub-section~\ref{subsec:attacks}. In the second, we check whether a classifier labels the adversarial reconstructions with the target shape class or if the source class label is avoided. In the third, we check our attacks' geometric measures with AEs other than the victim one.

\medskip
\noindent {\bfseries Geometric performance} \quad 
Table~\ref{tbl:attacks_loss_dist_chamfer} presents the adversarial metrics for latent and output attacks, where the results are averaged over the source-target pairs of each attack. Figure~\ref{fig:attack} shows visualizations. The latent space attack results in 25 $OS$ points, which is only about 1\% of the point set's total number of points. The attack increases the AE's reconstruction error for the target shape by a factor of 2.16. The output space attack reduces the $\TNRE$ to 1.11, which is higher than the AE's reference error by just 11\%.

The advantage of the latent space attack is that it does not need to know the decoder part of the network, and it back-propagates fewer gradients, only through the encoder part. However, this is also its drawback. The attack is limited, as it aims to produce the target's latent vector and is unfamiliar with the decoding process that follows. On the other hand, the output space attack has to know the entire AE model and propagates gradients from its end. However, it optimizes the target reconstruction directly, at the output of the AE. We conclude that this attack is successful in the geometric sense: a small perturbation to the source results in the desired output geometry.

\begin{figure}[tb!]
\begin{center}
\includegraphics[width=0.9\linewidth]{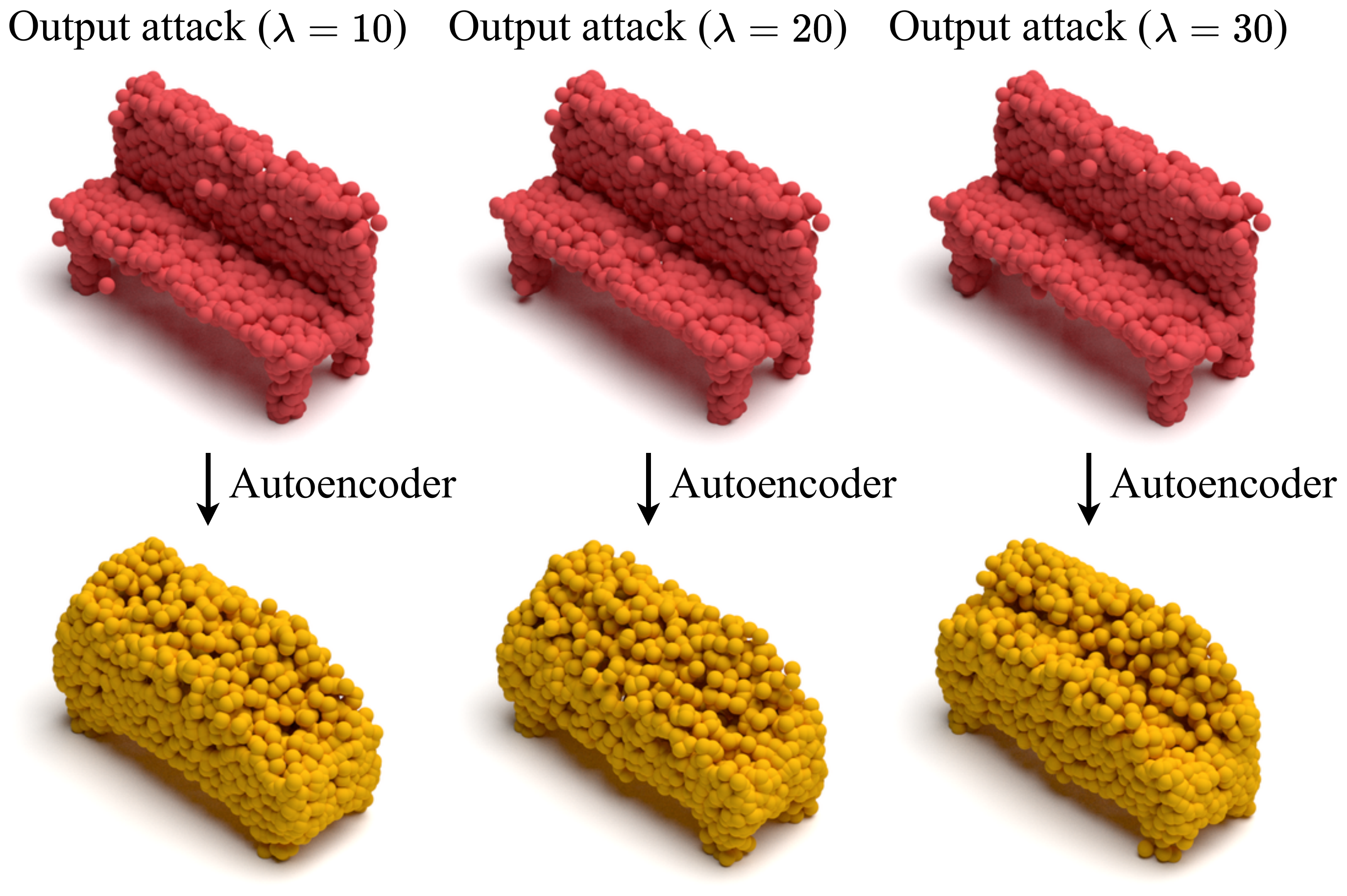}
\caption{{\bfseries Attack trade-off.} Top row: output space attack with different regularization strength ($\lambda$ in Equation~\ref{eq:problem_def_output}). Bottom row: corresponding reconstructions. A higher $\lambda$ reduces the input distortion and increases the reconstruction error. Still, even with the decreased perturbation, our attack can change the output geometry.}
\label{fig:attack_tradeoff}
\end{center}
\end{figure}

The latent and output space attacks offer a trade-off between the source distortion level and target reconstruction error: an increased regularization strength $\lambda$ reduces the former while raising the latter. This trade-off enables an attacker to select its preferred operation mode. In our study, we chose a $\lambda$ value that balances between the source and target metrics. Results for different $\lambda$ values for the example from Figure~\ref{fig:teaser} are shown in Figure~\ref{fig:attack_tradeoff}.

\begin{table}[tb!]
\centering
\begin{tabular}{@{ } l c c c c @{ }}
\toprule
Attack Type   & \#$OS$  & $\SCD$ & $\TRE$ & $\TNRE$  \\
\midrule
Latent attack & 25 & 0.53 & 1.40 & 2.16 \\
Output attack & {\bfseries 24} & {\bfseries 0.44} & {\bfseries 0.63} & {\bfseries 1.11} \\
\bottomrule
\end{tabular}
\vspace{0.2cm}
\caption{{\bfseries Geometric adversarial metrics.} The attacks are evaluated with the following metrics (a lower value is better). $OS$ (Equation ~\ref{eq:off_surface}) counts the number of off-surface points seen in an adversarial example. $\SCD=CD(Q,S)$ denotes Chamfer Distance between adversarial point cloud $Q$ and clean \textit{source} $S$, while $\TRE=CD(\widehat{Q}, T)$ is the Chamfer Distance between reconstructed adversarial set $\widehat{Q}$ and clean \textit{target} $T$. $\SCD$ and $\TRE$ are multiplied by a factor of $10^3$. $\TNRE$ (Equation ~\ref{eq:t_nre}) is the target normalized reconstruction error, relative to the AE's inherent error. The output space attack is better than the latent space attack in terms of both source distortion and target reconstruction quality.}
\label{tbl:attacks_loss_dist_chamfer}
\end{table}

\medskip
\noindent {\bfseries Semantic interpretation} \quad
The semantic \newtxt{by-product effect} of our attacks is evaluated as follows. We employ the train set of the victim AE (sub-section~\ref{subsec:experimental_setup}) and train a PointNet~\cite{qi2017pointnet} classifier. We utilize this classifier to check the classification accuracy for reconstructed adversarial inputs, where the ground truth label is the target's shape class. We also measure whether the predicted label is different from that of the source shape. As a baseline, we measure the accuracy for reconstructed clean target point sets selected for the attack. Table~\ref{tbl:attack_semantics} summarizes the results.

\begin{table}[tb!]
\centering
\begin{tabular}{@{ } l c c @{ }}
\toprule
Input Type     & Hit Target & Avoid Source \\
\midrule
Clean target (w/o cls)  & 64.6\% & 92.0\% \\ 
\midrule
Latent attack (w/o cls) & 45.8\% & 82.3\% \\
Output attack (w/o cls) & \textbf{55.3\%} & \textbf{88.6\%} \\
\midrule
Clean target (w/ cls)  & 90.9\% & 98.1\% \\ 
\midrule
Latent attack (w/ cls) & 59.0\% & 86.3\% \\
Output attack (w/ cls) & \textbf{76.0\%} & \textbf{94.7\%} \\
\bottomrule
\end{tabular}
\vspace{0.2cm}
\caption{{\bfseries Semantic interpretation.} We report classification results for reconstructed point clouds of clean targets, the latent space attack, and the output space attack. We measure the accuracy of predicting the target shape class label (Hit Target) or predicting a label other than the source class (Avoid Source). Targets are chosen from geometric nearest neighbor instances (w/o cls case) or only from neighbors with correct label prediction (w/ cls case). For all cases, a higher value is better. The 64.6\% accuracy for clean targets in the first case is relatively low since the attack selects instances near the shape class boundary that are hard to recognize correctly. Including the semantic consideration for the target selection improves the results considerably. Please see the "Semantic interpretation" part in sub-section~\ref{subsec:attack_evaluation} for more details.}
\label{tbl:attack_semantics}
\end{table}




We first notice that the baseline accuracy for clean targets is only 64.6\%. This result sheds some light on the attack's mechanism. The shape from a target class is selected at the boundary of the class, such that it is close \textit{geometrically} to the source shape. These target instances confuse the classifier and result in the relatively low accuracy. Still, the reconstructed targets avoid the source label for 92.0\%.

The accuracy for the output space attack drops by just 9.3\%, from 64.6\% to 55.3\%, and the source label is avoided 88.6\% of the time. If we allow our attack to select only targets with correct semantic interpretation, the results are improved to 76.0\% and 94.7\%. Please see the supplementary for further discussion. To sum up, while our attack hits the target class label partially, in most cases, it misleads the classifier into a different shape class than that of the source.

As semantic attacks on classifiers are widely common in the literature, an interesting question is whether they are effective as a geometric attack on an AE. To examine this question, we used the classifier for semantic interpretation of our attacks and generated adversarial examples with the popular perturbation attack of Xiang \etal~\cite{xiang2019generating}, for the source point clouds used in our work. Then, we fed the attack’s results through the AE and evaluated the success rate for the reconstructed point sets. A visualization of an adversarial example and its reconstruction is provided in Figure~\ref{fig:attack}.

The semantic attack altered the classifier’s prediction to the target label with a success rate of 100\%. However, after passing through the AE, the geometry remained the same, and the success rate dropped to 1\% only. This experiment suggests that semantic adversarial examples are highly ineffective against the AE, while our attacks are much more effective in both the geometric and semantic aspects.

\begin{figure}[tb!]
\begin{center}
\includegraphics[width=\columnwidth]{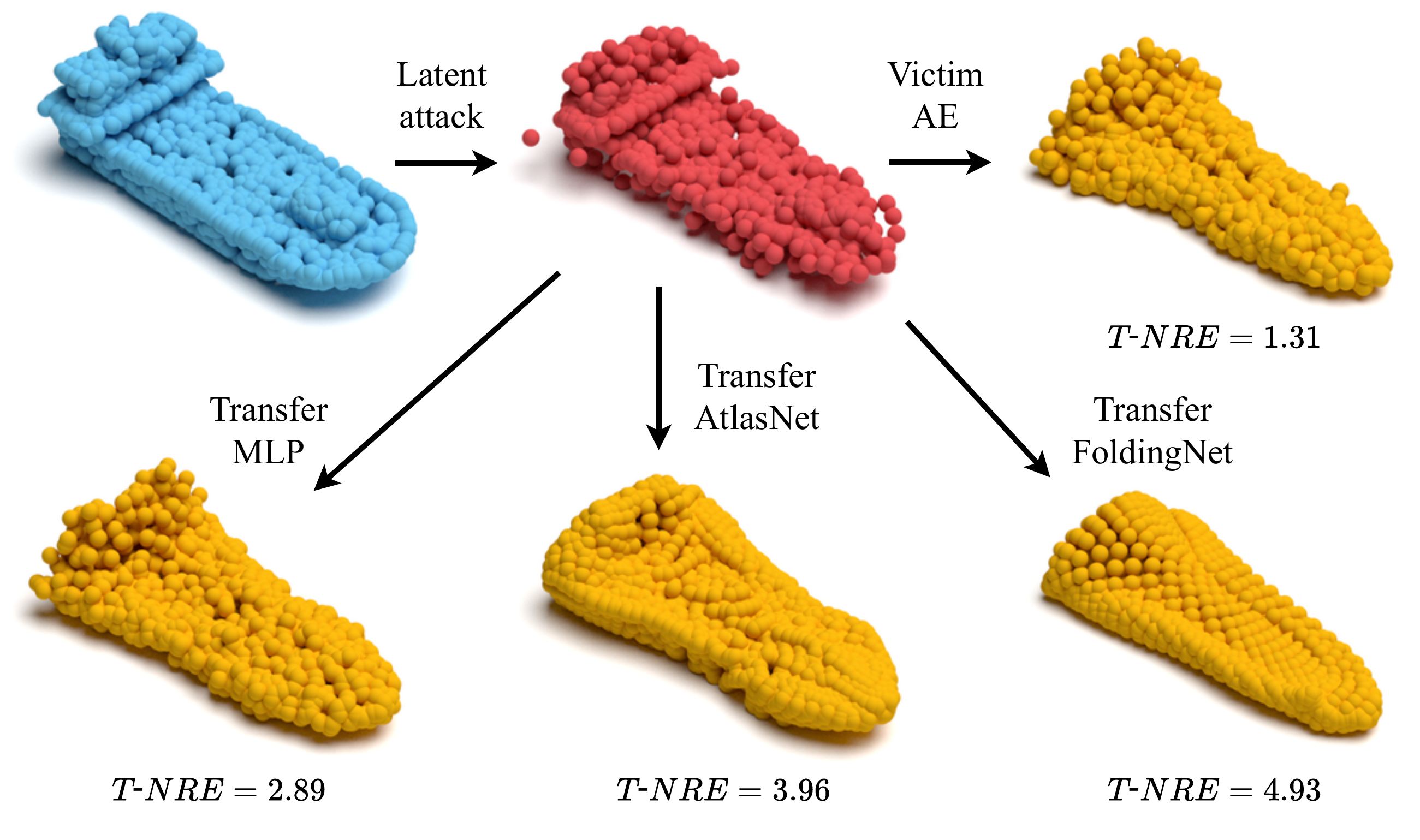}
\caption{{\bfseries Attack transfer.} Top row: a clean source (\textit{watercraft}), an adversarial example (targeted to an \textit{airplane}), and the reconstructed point cloud by the victim AE~\cite{achlioptas2018learning}. In the bottom row, reconstructions by other AEs: an AE with the same architecture as the attacked one and different initialization (MLP), AtlasNet~\cite{groueix2018atlasnet}, and FoldingNet~\cite{yang2018folding}. Here, the transfer to other AEs results in hybrids of airplane and watercraft shapes, and an increased $\TNRE$.}
\vspace{-10pt}
\label{fig:transfer}
\end{center}
\end{figure}

\medskip
\noindent {\bfseries Transferability} \quad
A typical test for an adversarial attack is whether it can be successfully transferred  to a different model than the one it was designed for. We evaluate the transferability of our attacks by feeding their adversarial examples through different AEs. One AE has the same MLP architecture as the attacked AE~\cite{achlioptas2018learning} but different random weight initialization. The other AEs are the widely used AtlaseNet~\cite{groueix2018atlasnet} and FoldingNet~\cite{yang2018folding}. We measure transferability with geometric metrics: the reconstruction error with respect to the \textit{target} point cloud and its normalization by the victim AE's original error (Equation~\ref{eq:t_nre}). Results are reported in Table~\ref{tbl:transfer}, and Figure~\ref{fig:transfer} shows an example.

The examples of the latent space attack transfer better to other AEs than the output space attack examples. As the former is optimized through only the encoder part of the victim AE, it is less coupled with that AE and thus more transferable. Still, in both cases, the error is higher compared to that of the attacked AE (Table~\ref{tbl:attacks_loss_dist_chamfer}).




\begin{table}[tb!]
\centering
\begin{tabular}{@{ } l c c @{ }}
\toprule
Transfer Type & $\TRE$ & $\TNRE$ \\
\midrule
Latent (MLP~\cite{achlioptas2018learning}, different init) & \textbf{3.65} & \textbf{5.70} \\
Output (MLP~\cite{achlioptas2018learning}, different init) & 4.26          & 7.51          \\
\midrule
Latent (AtlasNet~\cite{groueix2018atlasnet}) & \textbf{5.03} & \textbf{7.90} \\
Output (AtlasNet~\cite{groueix2018atlasnet}) & 7.60          & 13.23         \\
\midrule
Latent (FoldingNet~\cite{yang2018folding}) & \textbf{8.01} & \textbf{12.88} \\
Output (FoldingNet~\cite{yang2018folding}) & 10.10          & 17.39         \\
\bottomrule
\end{tabular}
\vspace{0.2cm}
\caption{{\bfseries Geometric attack transfer metrics.} Adversarial examples are reconstructed by different AEs. See the acronyms' definition in Table~\ref{tbl:attacks_loss_dist_chamfer}. Init stands for initialization. $\TRE$ is multiplied by a factor of $10^3$. A lower value is better. The transfer to a different AE results in high reconstruction error with respect to the attack's \textit{target} point cloud.}
\label{tbl:transfer}
\end{table}

The error increase is since, in the transfer setting, the reconstructions are a mix of the source and target shapes, as presented in Figure~\ref{fig:transfer}. This behavior may be due to the typical nature of an AE: it mostly captures the essence of the input rather than its peculiarities. We conclude that our attack is efficient when the model is known and degrades when it is not accessible to the adversary.




\medskip
\noindent {\bfseries Ablation study} \quad
\newtxt{We validated the design choices in our attack by a thorough ablation study and found that selecting targets according to their Chamfer Distance from the source contributes the most to the attack's success. It highlights the importance of this key geometric aspect in our method. Additional details are provided in the supplementary material.}





\subsection{Evaluation in Defense Setting}  \label{sec:defense_evaluation}
Similar to the adversarial evaluation, we use here geometric (sub-section~\ref{subsec:defenses}, "Defense metrics") and semantic (sub-section~\ref{subsec:attack_evaluation}, "Semantic interpretation") measures. Unlike the attack, the metrics are computed with respect to the \textit{source} rather than the \textit{target} point set. In the semantic evaluation, we measure whether the reconstructed shape of the defended input is labeled as the source class. 

\begin{figure*}[tb!]
\begin{center}
\includegraphics[width=0.7\linewidth]{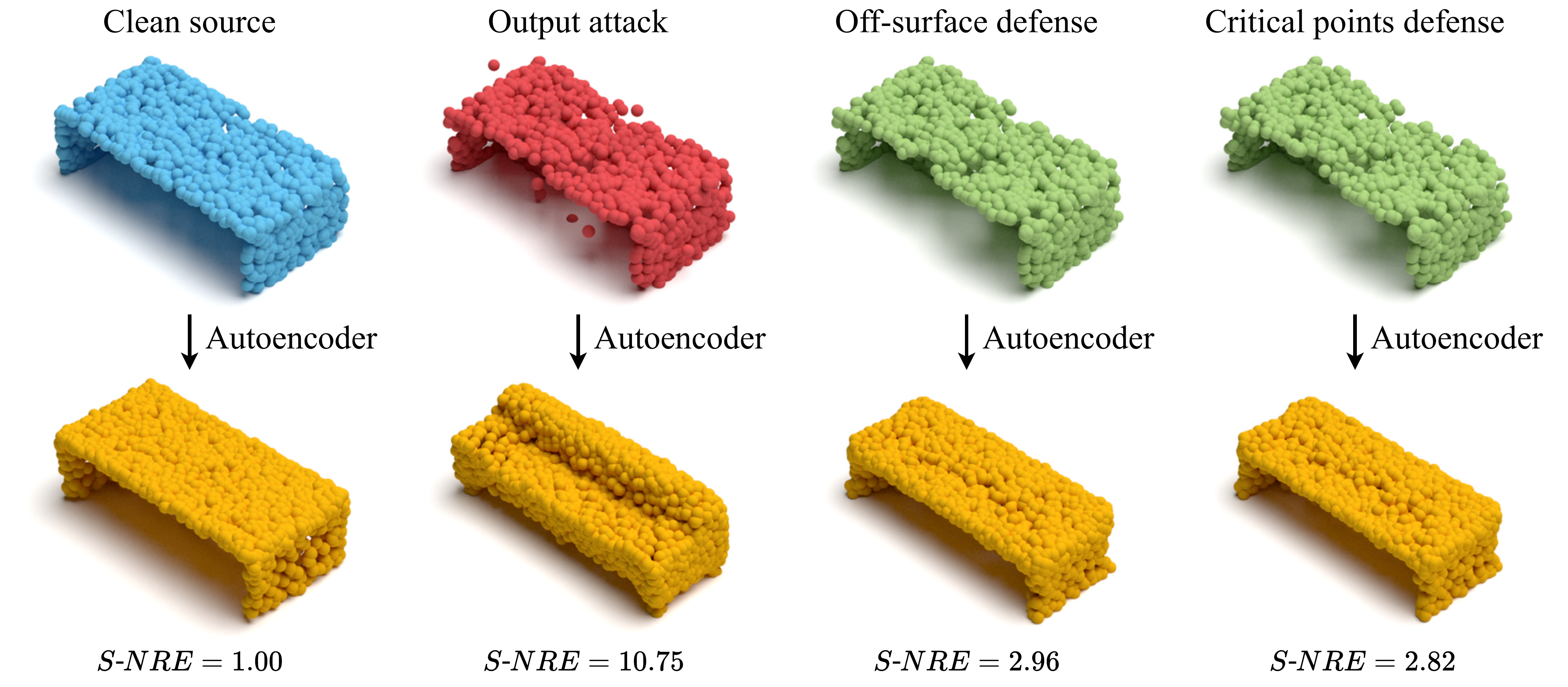}
\caption{{\bfseries Off-surface and critical points defenses.} Top row: a clean source point cloud (\textit{table}), an adversarial example of the output space attack (targeted to a \textit{sofa}), defended point cloud by the off-surface defense, and defended point cloud by the critical points defense. Bottom row: corresponding reconstructions and their source normalized reconstruction error, $\SNRE$, before defense (two results on the left) and after defense (two results on the right). While both defenses remove offensive points from the input, remnants of the attack's target still appear in the output reconstruction. Here, instead of a flat surface, the top of the reconstructed table after defense resembles the sofa's seat.}
\vspace{-3pt}
\label{fig:defense}
\end{center}
\end{figure*}


\medskip
\noindent {\bfseries Geometric performance} \quad
The $\SRE$ and $\SNRE$ before and after defense, for off-surface and critical points removal, are detailed in Table~\ref{tbl:defenses}. We report the defense's effect on clean source point clouds and adversarial examples of latent and output space attacks. Figure~\ref{fig:defense} presents a visualization of defended inputs and their reconstructions.

Both defenses have a negligible influence on the reconstruction of a defended clean source. As it has on-surface points, the error remains the same after the off-surface defense. When critical points are removed, other points in their vicinity become critical, and the error is increased by only 2\%. For an adversarial input, the defenses reduce the attack's effect, as reflected by the lower $\SRE$ and $\SNRE$. However, the geometric error is still high.

\begin{table}[tb!]
\centering
\begin{tabular}{@{ } l c c c @{ }}
\toprule
Defense Type    &  $\SRE$ $\downarrow$     & $\SNRE$ $\downarrow$ & $\SRCA$ $\uparrow$ \\
\midrule
Surface (Src)   &  0.58/\textbf{0.58} &  1.00/\textbf{1.00} & 91.7\%/\textbf{91.7\%} \\
Critical (Src)  &  0.58/0.59          &  1.00/1.02          & 91.7\%/91.4\%        \\
\midrule
Surface (Lat)  & 7.19/1.92          & 18.64/3.68          & 17.7\%/72.0\%          \\
Critical (Lat) & 7.19/\textbf{1.87} & 18.64/\textbf{3.66} & 17.7\%/\textbf{74.5\%} \\
\midrule
Surface (Out)  &  9.64/1.73          & 24.67/3.10          & 11.4\%/79.5\%          \\
Critical (Out) &  9.64/\textbf{1.68} & 24.67/\textbf{3.04} & 11.4\%/\textbf{80.7\%} \\
\bottomrule
\end{tabular}
\vspace{0.2cm}
\caption{{\bfseries Geometric defense metrics and semantics before/after the defense.} Off-surface or critical points defense is applied to both latent (Lat) and output (Out) space attacks. For comparison, we also include defense results for clean source shapes (Src). $\SRE$, $\SNRE$, and $\SRCA$ stand for source reconstruction error, source normalized reconstruction error, and source reconstruction classification accuracy. The first is in geometric distance units, the second is relative to the baseline error of the attacked AE, and the third is the percentage of correctly classified instances as the \textit{source} class label. $\SRE$ is multiplied by a factor of $10^3$. The arrows indicate whether a lower or higher value is better. The defenses reduce the reconstruction error with respect to the \textit{source} point cloud and increase the classification accuracy.}
\vspace{-1pt}
\label{tbl:defenses}
\end{table}

As Figure~\ref{fig:defense} shows, the output after defense is indeed more similar to the source than the adversarial output. Yet, we notice surprising behavior. Geometric features of the target shape bleed into the reconstruction of the defended point set. This phenomenon occurs since the attack not only moves points of the source to positions that cause the desired target reconstruction. It also perturbs points that prevent it. Thus, even after off-surface or critical points are removed, a residual effect of the attack is still present at the output, indicating the geometric robustness of our attack.




\medskip
\noindent {\bfseries Semantic interpretation} \quad
For semantic evaluation in the defense setting, we use the same classifier utilized for the adversarial evaluation, where the ground truth label now is that of the \textit{source} shape class. As reported in Table~\ref{tbl:defenses}, the reference classification accuracy for reconstructed clean source point clouds used for the attack is 91.7\%.

Before the defense, the accuracy of adversarial examples is very low - less than 18\%. However, applying the defense causes a considerable improvement. For instance, the critical points defense against the output space attack increases the accuracy to 80.7\%. This result is off by only 11\% from the 91.7\% reference accuracy for clean sources. \newtxt{We conclude that while the source's geometry is partially restored, its semantic meaning can be rescued.}


\section{Conclusions} \label{sec:conclusions}
\newtxt{In this paper, we presented \newtxt{for the first time} geometric adversarial attacks on 3D point clouds. Our attacks aim to craft adversarial examples that change the reconstructed geometry by an autoencoder model. It differs substantially from semantic attacks that operate on a classifier to manipulate its label prediction, as have been done before. We suggested two attack variants: a gray-box attack that targets a latent vector and a white-box attack, which optimizes the desired output shape directly and yields better reconstruction performance than the latent attack. We also examined the effect of deflecting off-surface or critical points from the adversarial examples on the attacks' performance.}



Our study revealed an interesting duality. The attack is highly successful geometrically. Meaning, a small adversarial perturbation to a clean point cloud leads to a different output shape with a low reconstruction error. However, it is interpreted to the target class label for about half of the examples. In contrast, the reconstructed defended point sets have a high geometric error with respect to the clean source, while their semantic interpretation is mostly restored.



\medskip
\noindent \textbf{Acknowledgment} \quad This work was partly funded by ISF grant number 1549/19.

{\small
\bibliographystyle{ieee}
\bibliography{references}

\begin{thebibliography}{10}\itemsep=-1pt

\bibitem{achlioptas2018learning}
P.~Achlioptas, O.~Diamanti, I.~Mitliagkas, and L.~J. Guibas.
\newblock {Learning Representations and Generative Models for 3D Point Clouds}.
\newblock In {\em Proceedings of the 35th International Conference on Machine
  Learning (ICML)}, pages 40--49, 2018.

\bibitem{cao2019adversarial}
Y.~Cao, C.~Xiao, D.~Yang, J.~Fang, R.~Yang, M.~Liu, and B.~Li.
\newblock {Adversarial Objects Against LiDAR-Based Autonomous Driving Systems}.
\newblock {\em arXiv preprint arXiv:1907.05418}, 2019.

\bibitem{carlini2017towards}
N.~Carlini and D.~Wagner.
\newblock {Towards Evaluating the Robustness of Neural Networks}.
\newblock {\em IEEE Symposium on Security and Privacy (SP)}, pages 39--57,
  2017.

\bibitem{chang2015shapenet}
A.~X. Chang, T.~Funkhouser, L.~J. Guibas, P.~Hanrahan, Q.~Huang, Z.~Li,
  S.~Savarese, M.~Savva, S.~Song, H.~Su, J.~Xiao, L.~Yi, and F.~Yu.
\newblock {ShapeNet: An Information-Rich 3D Model Repository}.
\newblock {\em arXiv preprint arXiv:1512.03012}, 2015.

\bibitem{chen2019learning}
Z.~Chen and H.~Zhang.
\newblock {Learning Implicit Fields for Generative Shape Modeling}.
\newblock In {\em Proceedings of the IEEE/CVF Conference on Computer Vision and
  Pattern Recognition (CVPR)}, pages 5939--5948, 2019.

\bibitem{dong2020self}
X.~Dong, D.~Chen, H.~Zhou, G.~Hua, W.~Zhang, and N.~Yu.
\newblock {Self-Robust 3D Point Recognition via Gather-Vector Guidance}.
\newblock In {\em Proceedings of the IEEE/CVF Conference on Computer Vision and
  Pattern Recognition (CVPR)}, pages 11516--11524, 2020.

\bibitem{dovrat2019learning}
O.~Dovrat, I.~Lang, and S.~Avidan.
\newblock {Learning to Sample}.
\newblock In {\em Proceedings of the IEEE/CVF Conference on Computer Vision and
  Pattern Recognition (CVPR)}, pages 2760--2769, 2019.

\bibitem{haoqiang2017apointset}
H.~Fan, H.~Su, and L.~Guibas.
\newblock {A Point Set Generation Network for 3D Object Reconstruction from a
  Single Image}.
\newblock In {\em Proceedings of the IEEE Conference on Computer Vision and
  Pattern Recognition (CVPR)}, pages 605--603, 2017.

\bibitem{gao2019sdm-net}
L.~Gao, J.~Yang, T.~Wu, Y.-J. Yuan, H.~Fu, Y.-K. Lai, and H.~Zhang.
\newblock {SDM-NET: Deep Generative Network for Structured Deformable Mesh}.
\newblock {\em ACM Transactions on Graphics (TOG), SIGGRAPH Asia 2019},
  38(6):Article 243, 2019.

\bibitem{goodfellow2015explaining}
I.~J. Goodfellow, J.~Shlens, and C.~Szegedy.
\newblock {Explaining and Harnessing Adversarial Examples}.
\newblock In {\em Proceedings of the International Conference on Learning
  Representations (ICLR)}, 2015.

\bibitem{groueix2018atlasnet}
T.~Groueix, M.~Fisher, V.~G. Kim, B.~C. Russell, and M.~Aubry.
\newblock {AtlasNet: A Papier-M\^ach\'e Approach to Learning 3D Surface
  Generation}.
\newblock In {\em Proceedings of the IEEE Conference on Computer Vision and
  Pattern Recognition (CVPR)}, pages 216--224, 2018.

\bibitem{hamdi2020advpc}
A.~Hamdi, S.~Rojas, A.~Thabet, and B.~Ghanem.
\newblock {AdvPC: Transferable Adversarial Perturbations on 3D Point Clouds}.
\newblock In {\em Proceedings of the European Conference on Computer Vision
  (ECCV)}, pages 877--894, 2020.

\bibitem{hermosilla2019total}
P.~Hermosilla, T.~Ritschel, and T.~Ropinski.
\newblock Total denoising: Unsupervised learning of 3d point cloud cleaning.
\newblock In {\em Proceedings of the IEEE/CVF International Conference on
  Computer Vision (ICCV)}, pages 52--60, 2019.

\bibitem{hinton2006reducing}
G.~E. Hinton and R.~R. Salakhutdinov.
\newblock {Reducing the Dimensionality of Data with Neural Networks}.
\newblock {\em Science}, 313:504--507, 2006.

\bibitem{kingma2014auto}
D.~P. Kingma and M.~Welling.
\newblock {Auto-Encoding Variational Bayes}.
\newblock In {\em Proceedings of the International Conference on Learning
  Representations (ICLR)}, 2014.

\bibitem{landrieu2018large}
L.~Landrieu and M.~Simonovsky.
\newblock {Large-scale Point Cloud Semantic Segmentation with Superpoint
  Graphs}.
\newblock In {\em Proceedings of the IEEE Conference on Computer Vision and
  Pattern Recognition (CVPR)}, pages 4558--4567, 2018.

\bibitem{lang2020samplenet}
I.~Lang, A.~Manor, and S.~Avidan.
\newblock {SampleNet: Differentiable Point Cloud Sampling}.
\newblock In {\em Proceedings of the IEEE/CVF Conference on Computer Vision and
  Pattern Recognition (CVPR)}, pages 7578--7588, 2020.

\bibitem{lee2020shapeadv}
K.~Lee, Z.~Chen, X.~Yan, R.~Urtasun, and E.~Yumer.
\newblock {ShapeAdv: Generating Shape-Aware Adversarial 3D Point Clouds}.
\newblock {\em arXiv preprint arXiv:2005.11626}, 2020.

\bibitem{li2018so-net}
J.~Li, B.~M. Chen, and G.~H. Lee.
\newblock {SO-Net: Self-Organizing Network for Point Cloud Analysis}.
\newblock In {\em Proceedings of the IEEE Conference on Computer Vision and
  Pattern Recognition (CVPR)}, pages 9397--9406, 2018.

\bibitem{li2019pu-gan}
R.~Li, X.~Li, C.-W. Fu, D.~Cohen-Or, and P.-A. Heng.
\newblock {PU-GAN: a Point Cloud Upsampling Adversarial Network}.
\newblock In {\em Proceedings of the IEEE/CVF International Conference on
  Computer Vision (ICCV)}, pages 7203--7212, 2019.

\bibitem{liu2019extending}
D.~Liu, R.~Yu, and H.~Su.
\newblock {Extending Adversarial Attacks and Defenses to Deep 3D Point Cloud
  Classifiers}.
\newblock In {\em Proceedings of the IEEE International Conference on Image
  Processing (ICIP)}, 2019.

\bibitem{mehr2019disconet}
E.~Mehr, A.~Jourdan, N.~Thome, M.~Cord, and V.~Guitteny.
\newblock {DiscoNet: Shapes Learning on Disconnected Manifolds for 3D Editing}.
\newblock In {\em Proceedings of the IEEE/CVF International Conference on
  Computer Vision (ICCV)}, pages 3474--3483, 2019.

\bibitem{mescheder2019occupancy}
L.~Mescheder, M.~Oechsle, M.~Niemeyer, S.~Nowozin, and A.~Geiger.
\newblock {Occupancy Networks: Learning 3D Reconstruction in Function Space}.
\newblock In {\em Proceedings of the IEEE/CVF Conference on Computer Vision and
  Pattern Recognition (CVPR)}, pages 4460--4470, 2019.

\bibitem{mo2019structurenet}
K.~Mo, P.~Guerrero, L.~Yi, H.~Su, P.~Wonka, N.~Mitra, and L.~Guibas.
\newblock {StructureNet: Hierarchical Graph Networks for 3D Shape Generation}.
\newblock {\em ACM Transactions on Graphics (TOG), SIGGRAPH Asia 2019},
  38(6):Article 242, 2019.

\bibitem{mo2020structedit}
K.~Mo, P.~Guerrero, L.~Yi, H.~Su, P.~Wonka, N.~J. Mitra, and L.~Guibas.
\newblock {StructEdit: Learning Structural Shape Variations}.
\newblock In {\em Proceedings of the IEEE/CVF Conference on Computer Vision and
  Pattern Recognition (CVPR)}, pages 8859--8868, 2020.

\bibitem{papernot2016the}
N.~Papernot, P.~McDaniel, S.~Jha, M.~Fredrikson, Z.~B. Celik, and A.~Swami.
\newblock {The Limitations of Deep Learning in Adversarial Settings}.
\newblock {\em IEEE European Symposium on Security and Privacy (EuroS\&P)},
  pages 372--387, 2016.

\bibitem{qi2019deep}
C.~R. Qi, O.~Litany, K.~He, and L.~J. Guibas.
\newblock {Deep Hough Voting for 3D Object Detection in Point Clouds}.
\newblock {\em Proceedings of the International Conference on Computer Vision
  (ICCV)}, pages 9277--9286, 2019.

\bibitem{qi2017pointnet}
C.~R. Qi, H.~Su, K.~Mo, and L.~J. Guibas.
\newblock {PointNet: Deep Learning on Point Sets for 3D Classification and
  Segmentation}.
\newblock In {\em Proceedings of the IEEE Conference on Computer Vision and
  Pattern Recognition (CVPR)}, pages 652--660, 2017.

\bibitem{qi2017pointnetpp}
C.~R. Qi, L.~Yi, H.~Su, and L.~J. Guibas.
\newblock {PointNet++: Deep Hierarchical Feature Learning on Point Sets in a
  Metric Space}.
\newblock In {\em Proceedings of Advances in Neural Information Processing
  Systems (NeuralIPS)}, pages 5099--5108, 2017.

\bibitem{rakotosaona2020pointcleannet}
M.-J. Rakotosaona, V.~La~Barbera, P.~Guerrero, N.~J. Mitra, and M.~Ovsjanikov.
\newblock {PointCleanNet: Learning to Denoise and Remove Outliers from Dense
  Point Clouds}.
\newblock {\em Computer Graphics Forum}, 39(1):185--203, 2020.

\bibitem{shi2019pointrcnn}
S.~Shi, X.~Wang, and H.~Li.
\newblock {PointRCNN: 3D Object Proposal Generation and Detection From Point
  Cloud}.
\newblock In {\em The IEEE Conference on Computer Vision and Pattern
  Recognition (CVPR)}, pages 770--779, 2019.

\bibitem{su2019one}
J.~Su, D.~V. Vargas, and S.~Kouichi.
\newblock {One Pixel Attack for Fooling Deep Neural Networks}.
\newblock {\em IEEE Transactions on Evolutionary Computation}, 23(5):828--841,
  2019.

\bibitem{sun2020towards}
J.~Sun, Y.~Cao, Q.~A. Chen, and Z.~M. Mao.
\newblock {Towards Robust LiDAR-based Perception in Autonomous Driving: General
  Black-box Adversarial Sensor Attack and Countermeasures}.
\newblock In {\em Proceedings of the 29th USENIX Security Symposium}, 2020.

\bibitem{tchapmi2017segcloud}
L.~P. Tchapmi, C.~B. Choy, I.~Armeni, J.~Gwak, and S.~Savarese.
\newblock {SEGCloud: Semantic Segmentation of 3D Point Clouds}.
\newblock In {\em Proceedings of the International Conference on 3D Vision
  (3DV)}, 2017.

\bibitem{tu2020physically}
J.~Tu, M.~Ren, S.~Manivasagam, M.~Liang, B.~Yang, R.~Du, F.~Cheng, and
  R.~Urtasun.
\newblock {Physically Realizable Adversarial Examples for LiDAR Object
  Detection}.
\newblock In {\em Proceedings of the IEEE/CVF Conference on Computer Vision and
  Pattern Recognition (CVPR)}, pages 13716--13725, 2020.

\bibitem{wang2019graph}
L.~Wang, Y.~Huang, Y.~Hou, S.~Zhang, and J.~Shan.
\newblock {Graph Attention Convolution for Point Cloud Semantic Segmentation}.
\newblock In {\em Proceedings of the IEEE/CVF Conference on Computer Vision and
  Pattern Recognition (CVPR)}, pages 10296--10305, 2019.

\bibitem{wang2020cascaded}
X.~Wang, M.~H. A.~J. ~, and G.~H. Lee.
\newblock Cascaded refinement network for point cloud completion.
\newblock In {\em Proceedings of the IEEE/CVF Conference on Computer Vision and
  Pattern Recognition (CVPR)}, pages 790--799, 2020.

\bibitem{wang2019dynamic}
Y.~Wang, Y.~Sun, Z.~Liu, S.~E. Sarma, M.~M. Bronstein, and J.~M. Solomon.
\newblock {Dynamic Graph CNN for Learning on Point Clouds}.
\newblock {\em ACM Transactions on Graphics (TOG)}, 2019.

\bibitem{wen2020geometry}
Y.~Wen, J.~Lin, K.~Chen, C.~L.~P. Chen, and K.~Jia.
\newblock {Geometry-Aware Generation of Adversarial Point Clouds}.
\newblock {\em IEEE Transactions on Pattern Analysis and Machine Intelligence},
  2020.

\bibitem{wu2015modelnet}
Z.~Wu, S.~Song, A.~Khosla, F.~Yu, L.~Zhang, X.~Tang, and J.~Xiao.
\newblock {3D ShapeNets: A Deep Representation for Volumetric Shapes}.
\newblock {\em Proceedings of the IEEE Conference on Computer Vision and
  Pattern Recognition (CVPR)}, pages 1912--1920, 2015.

\bibitem{xiang2019generating}
C.~Xiang, C.~R. Qi, and B.~Li.
\newblock {Generating 3D Adversarial Point Clouds}.
\newblock In {\em Proceedings of the IEEE/CVF Conference on Computer Vision and
  Pattern Recognition (CVPR)}, pages 9136--9144, 2019.

\bibitem{yang2019pointflow}
G.~Yang, X.~Huang, Z.~Hao, M.-Y. Liu, S.~Belongie, and B.~Hariharan.
\newblock {PointFlow: 3D Point Cloud Generation With Continuous Normalizing
  Flows}.
\newblock In {\em Proceedings of the IEEE/CVF International Conference on
  Computer Vision (ICCV)}, pages 4541--4550, 2019.

\bibitem{zhang2019adversarial}
J.~Yang, Q.~Zhang, R.~Fang, B.~Ni, J.~Liu, and T.~Qi.
\newblock {Adversarial Attack and Defense on Point Sets}.
\newblock {\em arXiv preprint arXiv:1902.10899}, 2019.

\bibitem{yang2018folding}
Y.~Yang, C.~Feng, Y.~Shen, and D.~Tian.
\newblock {FoldingNet: Point Cloud Auto-encoder via Deep Grid Deformation}.
\newblock In {\em Proceedings of the IEEE Conference on Computer Vision and
  Pattern Recognition (CVPR)}, pages 206--215, 2018.

\bibitem{yang20203dssd}
Z.~Yang, Y.~Sun, S.~Liu, and J.~Jia.
\newblock {3DSSD: Point-Based 3D Single Stage Object Detector}.
\newblock In {\em Proceedings of the IEEE/CVF Conference on Computer Vision and
  Pattern Recognition (CVPR)}, pages 11040--11048, 2020.

\bibitem{yu2018pu-net}
L.~Yu, X.~Li, C.-W. Fu, D.~Cohen-Or, and P.~A. Heng.
\newblock {PU-Net: Point Cloud Upsampling Network}.
\newblock In {\em Proceedings of the IEEE Conference on Computer Vision and
  Pattern Recognition (CVPR)}, pages 2790--2799, 2018.

\bibitem{zhao20193d}
Y.~Zhao, T.~Birdal, H.~Deng, and F.~Tombari.
\newblock {3D Point Capsule Networks}.
\newblock In {\em Proceedings of the IEEE/CVF Conference on Computer Vision and
  Pattern Recognition (CVPR)}, pages 1009--1018, 2019.

\bibitem{zhao2020on}
Y.~Zhao, Y.~Wu, C.~Chen, and A.~Lim.
\newblock {On Isometry Robustness of Deep 3D Point Cloud Models Under
  Adversarial Attacks}.
\newblock In {\em Proceedings of the IEEE/CVF Conference on Computer Vision and
  Pattern Recognition (CVPR)}, pages 1201--1210, 2020.

\bibitem{zheng2019pointcloud}
T.~Zheng, C.~Chen, J.~Yuan, B.~Li, and K.~Ren.
\newblock {PointCloud Saliency Maps}.
\newblock In {\em Proceedings of the IEEE/CVF International Conference on
  Computer Vision (ICCV)}, pages 1961--1970, 2019.

\bibitem{zhou2020lg-gan}
H.~Zhou, D.~Chen, J.~Liao, K.~Chen, X.~Dong, K.~Liu, W.~Zhang, G.~Hua, and
  N.~Yu.
\newblock {LG-GAN: Label Guided Adversarial Network for Flexible Targeted
  Attack of Point Cloud Based Deep Networks}.
\newblock In {\em Proceedings of the IEEE/CVF Conference on Computer Vision and
  Pattern Recognition (CVPR)}, pages 10356--10365, 2020.

\bibitem{zhou2019dup-net}
H.~Zhou, K.~Chen, W.~Zhang, H.~Fang, W.~Zhou, and N.~Yu.
\newblock {DUP-Net: Denoiser and Upsampler Network for 3D Adversarial Point
  Clouds Defense}.
\newblock In {\em Proceedings of the IEEE/CVF International Conference on
  Computer Vision (ICCV)}, pages 1961--1970, 2019.

\end{thebibliography}
}

\clearpage
\appendix

\section*{Supplementary Material}
In the following sections, we provide more information regarding our geometric attacks and defenses. Sections~\ref{sec:supp_additional_attack} and~\ref{sec:supp_additional_defense} provide additional attack and defense results, respectively. An extensive ablation study is presented in Section~\ref{sec:supp_ablation_study}. In section~\ref{sec:supp_attack_modelnet}, we present attack results for another dataset~\cite{wu2015modelnet}. Finally, in Section~\ref{sec:supp_experimental_sett}, we detail experimental settings, including network architecture and optimization parameters for the victim autoencoder (AE), AEs for transfer, and the classifier for semantic evaluation.

\section{Additional Attack Results} \label{sec:supp_additional_attack}

\subsection{Evolution of the Attack} \label{sec:supp_evolution_attack}
\begin{figure*}[tb!]
\begin{center}
\begin{tabular}{ c c c c c }
\includegraphics[width=0.17\linewidth]{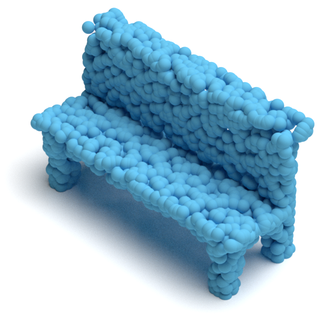} &
\includegraphics[width=0.17\linewidth]{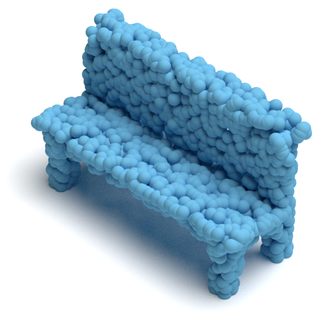} &
\includegraphics[width=0.17\linewidth]{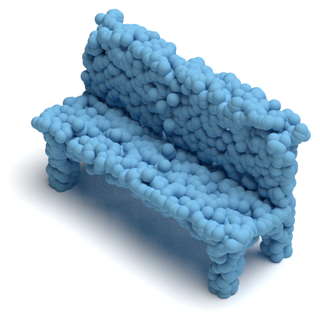} &
\includegraphics[width=0.17\linewidth]{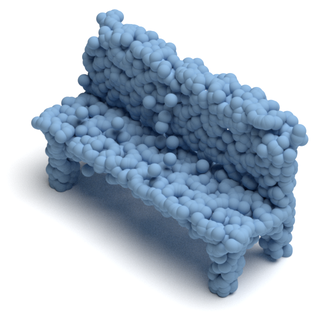} &
\includegraphics[width=0.17\linewidth]{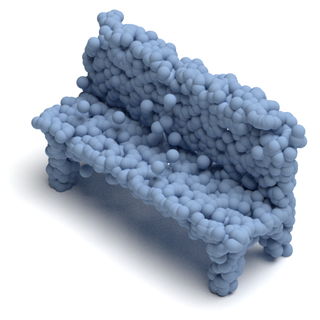} \\

\includegraphics[width=0.17\linewidth]{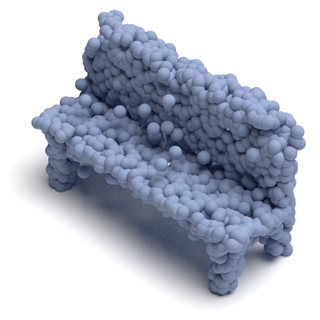} &
\includegraphics[width=0.17\linewidth]{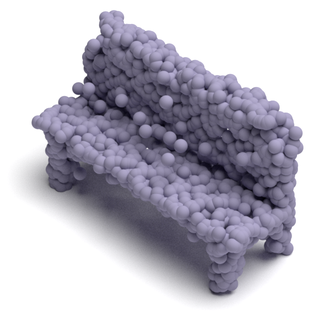} &
\includegraphics[width=0.17\linewidth]{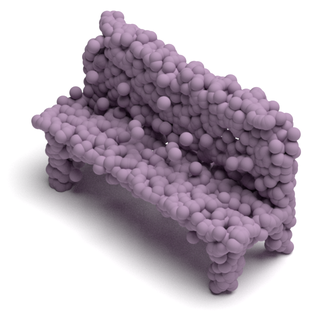} &
\includegraphics[width=0.17\linewidth]{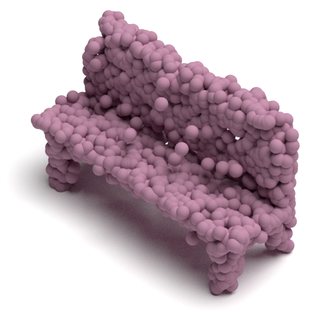} &
\includegraphics[width=0.17\linewidth]{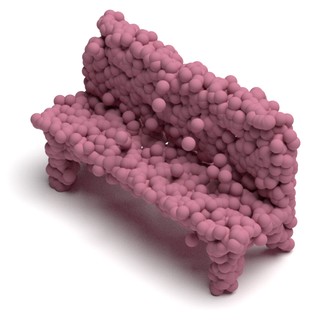} \\

\includegraphics[width=0.17\linewidth]{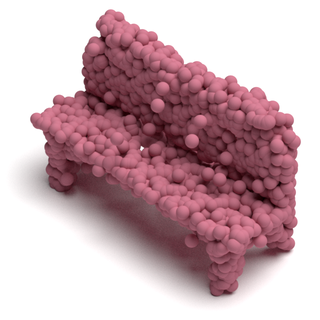} &
\includegraphics[width=0.17\linewidth]{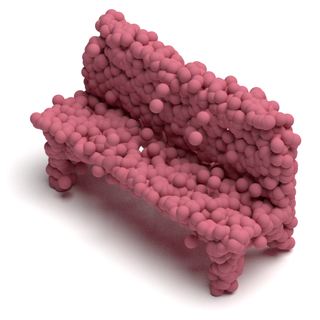} &
\includegraphics[width=0.17\linewidth]{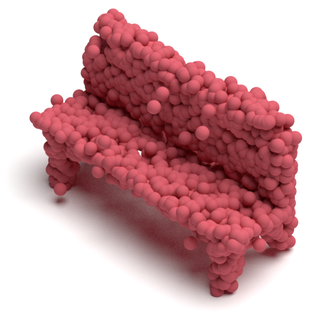} &
\includegraphics[width=0.17\linewidth]{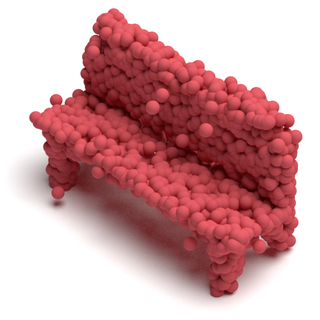} &
\includegraphics[width=0.17\linewidth]{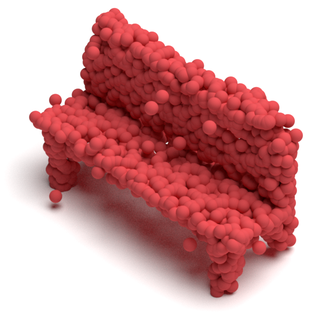} \\

\\

\includegraphics[width=0.17\linewidth]{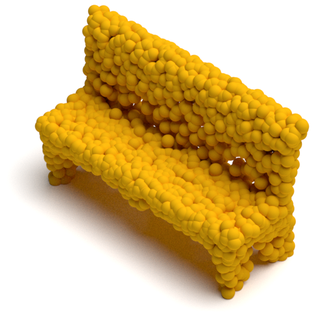} &
\includegraphics[width=0.17\linewidth]{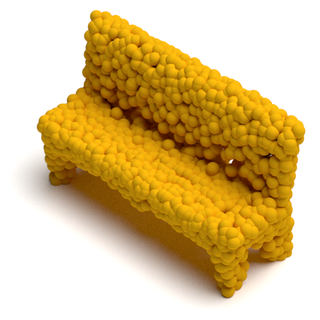} &
\includegraphics[width=0.17\linewidth]{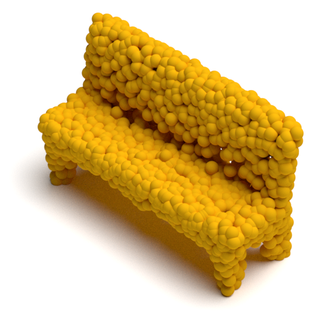} &
\includegraphics[width=0.17\linewidth]{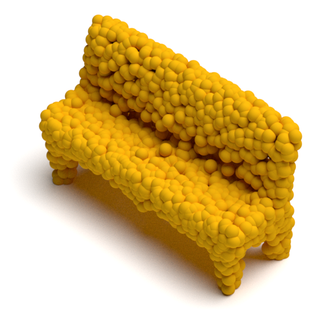} &
\includegraphics[width=0.17\linewidth]{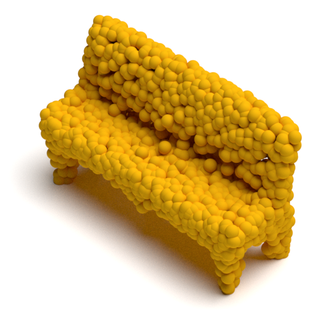} \\

\includegraphics[width=0.17\linewidth]{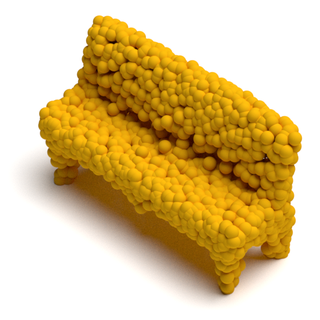} &
\includegraphics[width=0.17\linewidth]{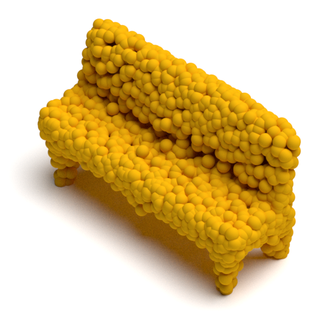} &
\includegraphics[width=0.17\linewidth]{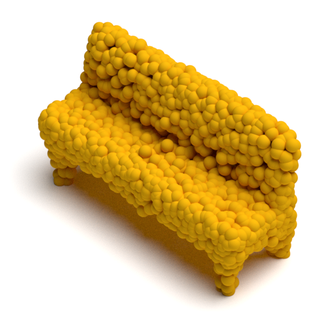} &
\includegraphics[width=0.17\linewidth]{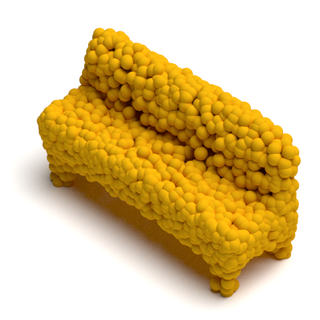} &
\includegraphics[width=0.17\linewidth]{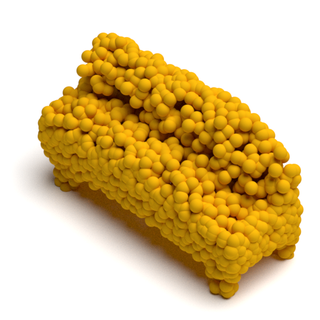} \\

\includegraphics[width=0.17\linewidth]{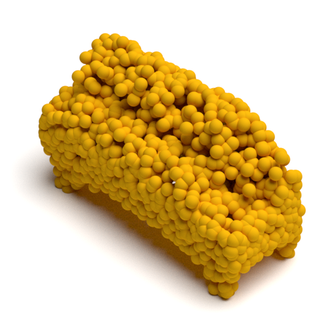} &
\includegraphics[width=0.17\linewidth]{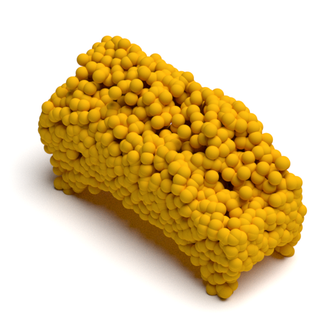} &
\includegraphics[width=0.17\linewidth]{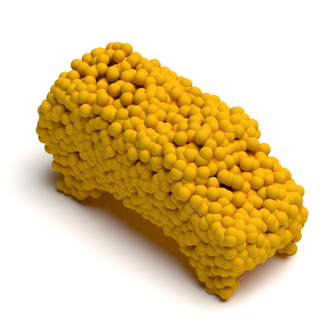} &
\includegraphics[width=0.17\linewidth]{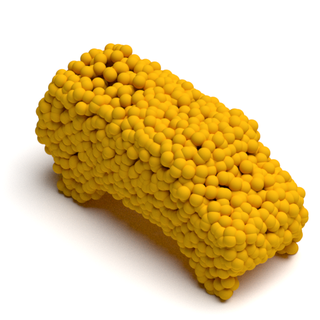} &
\includegraphics[width=0.17\linewidth]{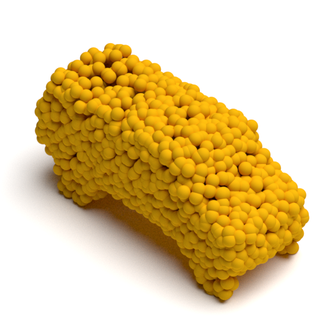} \\

\end{tabular}
\caption{{\bfseries Attack evolution.} Top three rows: interpolation between a clean source point cloud (a \textit{bench}, leftmost at the first row) and an adversarial example (targeted to a \textit{car}, rightmost at the third row). Bottom three rows: corresponding reconstructions. The interpolation weight $\alpha$ from left to right is: $\{0.0, 0.05, 0.1, 0.2, 0.25\}$ (first row), $\{0.3, 0.4, 0.5, 0.6, 0.7\}$ (second row), and $\{0.75, 0.8, 0.9, 0.95, 1.0 \}$ (third row). The adversarial point set evolves such that its reconstructions transition smoothly from the source to the target shape.}
\label{fig:supp_interp}
\end{center}
\end{figure*}

We simulate our attack's evolution by interpolating between a clean point set $S$ and a corresponding adversarial example $Q$. Specifically, we compute:
\begin{equation} \label{eq:supp_interp_input}
U = (1 - \alpha) S + \alpha Q,
\end{equation}

\noindent and:
\begin{equation} \label{eq:supp_interp_recon}
\widehat{U} = f_{AE}(U),
\end{equation}

\noindent where $\alpha \in [0, 1]$ is an interpolation factor, and $f_{AE}$ is the victim AE model. The process is shown in Figure~\ref{fig:supp_interp}, for the attack example in Figure~\ref{fig:teaser} in the paper.

The attack starts by exploring points to perturb (first row in Figure~\ref{fig:supp_interp}). Then, it realizes that some points are not required for imposing the target output shape. Thus, it shifts them towards the surface of the source shape to comply with the input distance loss (Equation~\ref{eq:loss_dist}) and continues to move the necessary points to satisfy the adversarial loss (Equation~\ref{eq:loss_output}). This process can be seen in the second and third rows of the figure. It implies that the attack converges to an adversarial example that maintains similarity to the source point cloud while achieving the desired target reconstruction.

\subsection{Untargeted Attack Setting} \label{subsec:supp_untargeted_attack_setting}
In addition to the targeted setting presented in the paper (sub-section~\ref{subsec:experimental_setup}), our attacks can be applied in an untargeted fashion: for each source instance, we select the best result over the target classes, in terms of the lowest attack score (Equation~\ref{eq:attack_score}). Thus, in the untargeted attack, there are $25 \cdot 13 = 325$ source-target pairs. Geometric metrics, averaged over the attack's pairs, are presented in Table~\ref{tbl:supp_attacks_untargeted}. As the optimal target shape is selected per source, the untargeted setting yields an improved geometric performance compared to the targeted attacks (Table~\ref{tbl:attacks_loss_dist_chamfer}).


\begin{table}[tb!]
\centering
\begin{tabular}{@{ } l c c c c @{ }}
\toprule
Attack Type   & \#$OS$  & $\SCD$ & $\TRE$ & $\TNRE$  \\
\midrule
Latent attack & 18 & 0.32 & 0.43 & 1.07 \\
Output attack & {\bfseries 11} & {\bfseries 0.12} & {\bfseries 0.35} & {\bfseries 0.91} \\
\bottomrule
\end{tabular}
\vspace{0.2cm}
\caption{{\bfseries Geometric adversarial metrics for untargeted attacks.} The acronyms and metrics are the same as in Table~\ref{tbl:attacks_loss_dist_chamfer}. $\SCD$ and $\TRE$ are multiplied by a factor of $10^3$. A lower value is better. In the untargeted setting, the attacks result in both small source distortion and low target reconstruction error.}
\label{tbl:supp_attacks_untargeted}
\end{table}

\subsection{Untargeted Attack Distribution} \label{subsec:supp_untargeted_attack_hist}
\begin{figure}[tb!]
\begin{center}
\includegraphics[width=\columnwidth]{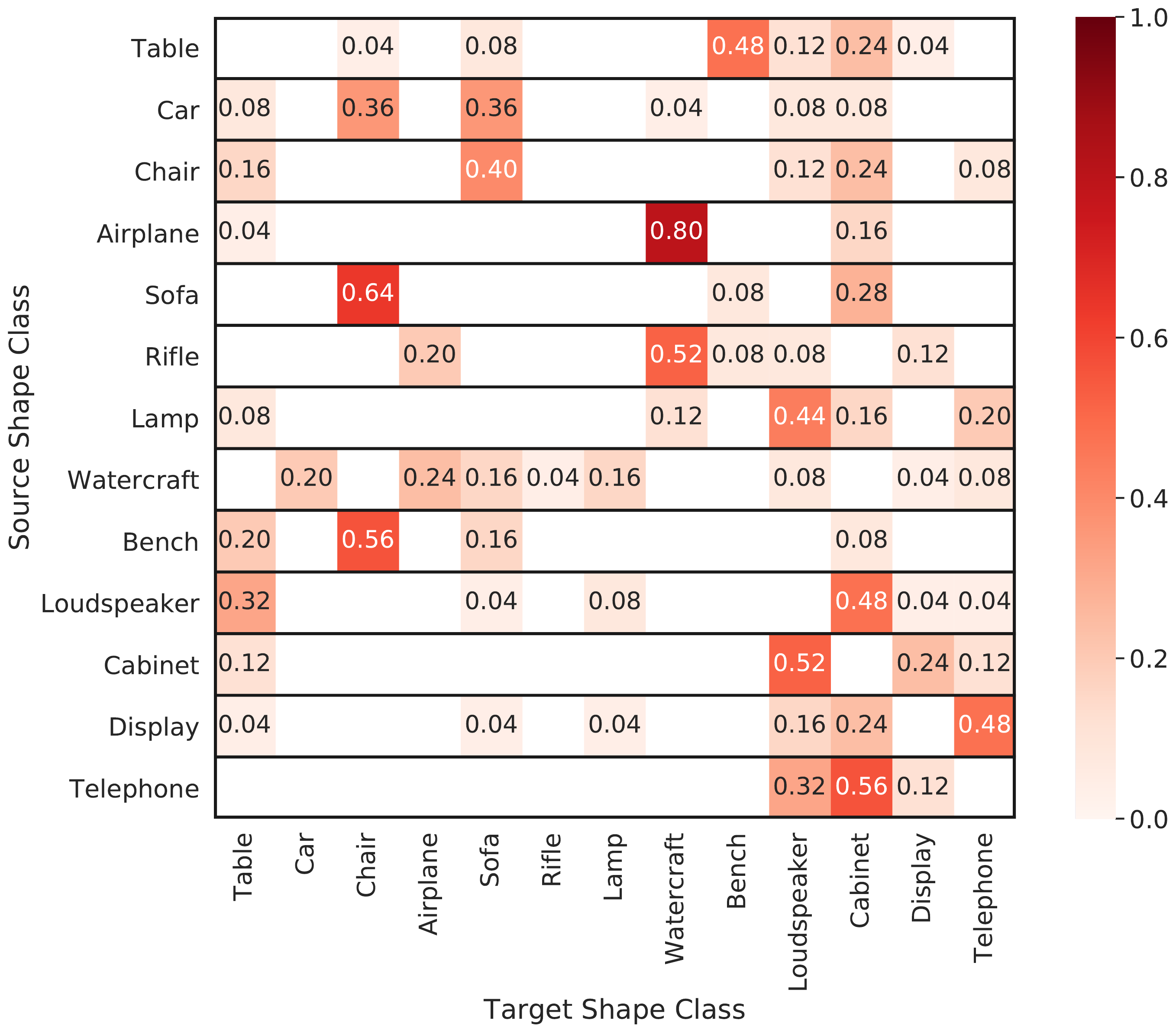}
\caption{{\bfseries  Untargeted output space attack distribution.} Each row presents the distribution of different target shape classes for the corresponding source shape class. For each source class, the distribution is mostly concentrated at one or two target classes.}
\label{fig:supp_untargeted_hist}
\end{center}
\end{figure}

To gain more insight into the relations between shape classes, we analyze the statistics of the untargeted output space attack. We compute the distribution over the selected target classes for the adversarial point clouds from each source class. Figure~\ref{fig:supp_untargeted_hist} presents the results. 

The figure points out that for each source class, there are a few favorable target classes. These classes contain instances that are geometrically close to the clean source point set. These instances enable the untargeted attack to craft adversarial examples with \textit{both} low input distortion \textit{and} low output reconstruction error, as Figure~\ref{fig:supp_untargeted_example} illustrates. In the rest of the supplemental material, we show results for the targeted attack setting unless mentioned otherwise.

\begin{figure}[tb!]
\begin{center}
\begin{tabular}{ c c c }
Clean  & Output space & Clean  \\
source & attack       & target \\

\includegraphics[width=0.28\linewidth]{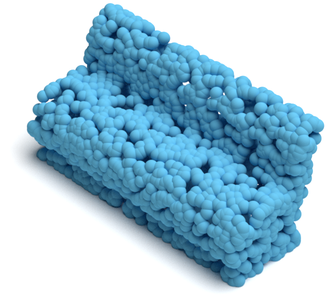} &
\includegraphics[width=0.28\linewidth]{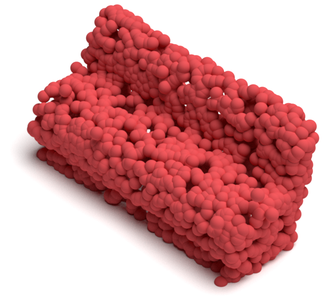} &
\includegraphics[width=0.28\linewidth]{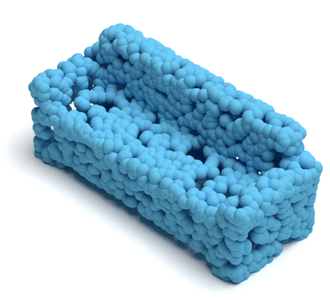} \\

\includegraphics[width=0.28\linewidth]{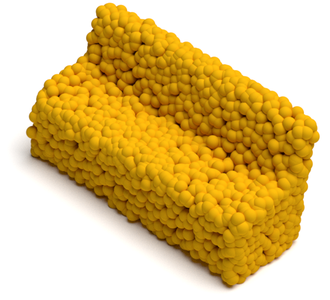} &
\includegraphics[width=0.28\linewidth]{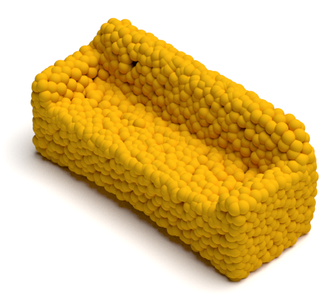} &
\includegraphics[width=0.28\linewidth]{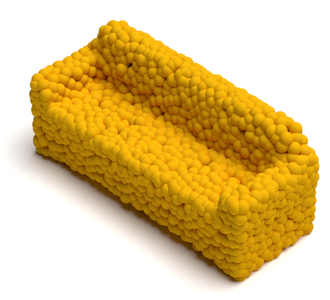} \\

N/A           & $\TNRE = 0.92$    & $\TNRE = 1.00$ \\
\end{tabular}
\caption{{\bfseries An untargeted attack example.} Top row: a clean source point cloud (\textit{bench}), an untargeted attack example, and a clean target (\textit{sofa}). Bottom row: corresponding reconstructions and their target normalized reconstruction error, $\TNRE$ (not applicable for the reconstructed source). In the untargeted setting, the adversarial perturbation is very small, and the reconstruction quality matches the reference reconstruction result for a clean target, as indicated by the low $\TNRE$ value.}
\label{fig:supp_untargeted_example}
\end{center}
\end{figure}

\subsection{Semantic Interpretation} \label{sec:supp_semantic_interp}
\begin{figure}[tb!]
\begin{center}
\includegraphics[width=\columnwidth]{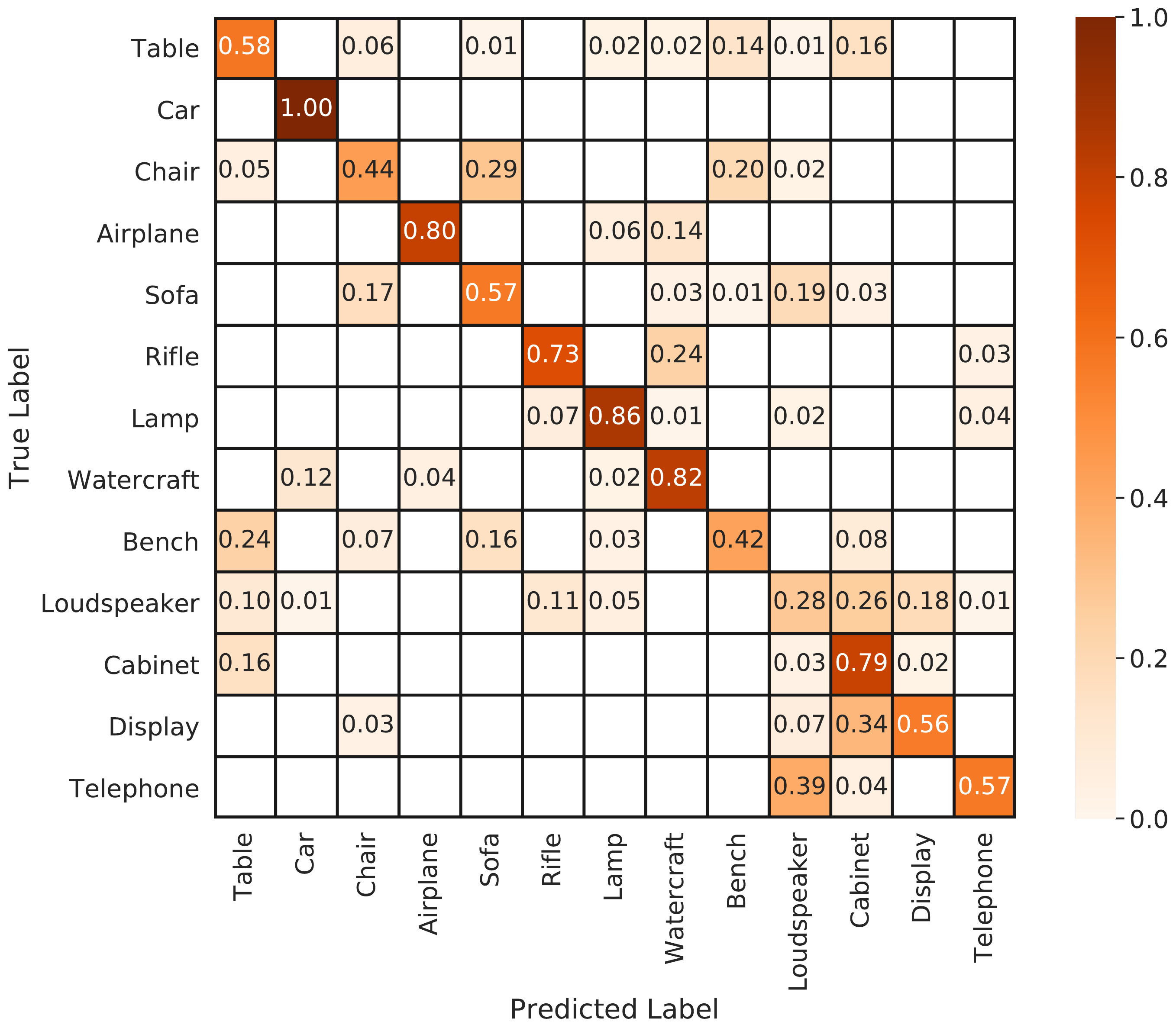}
\caption{{\bfseries Confusion matrix for reconstructed clean target point clouds.} Each cell shows the fraction of predicted class label (in the columns) for the corresponding true label (in the rows). Confusion occurs between shape classes that share similar geometric characteristics, such as \textit{chair} and \textit{sofa} classes.}
\label{fig:supp_confusion}
\end{center}
\end{figure}

\begin{figure}[tb!]
\begin{center}
\begin{tabular}{ c c c }
Clean  & Output space & Clean  \\
source & attack       & target \\
\includegraphics[width=0.28\linewidth]{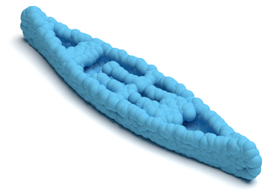} &
\includegraphics[width=0.28\linewidth]{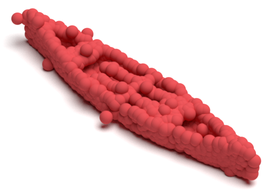} &
\includegraphics[width=0.28\linewidth]{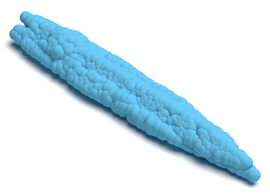} \\

\includegraphics[width=0.28\linewidth]{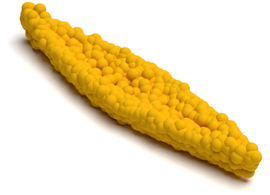} &
\includegraphics[width=0.28\linewidth]{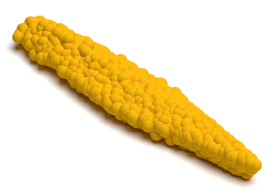} &
\includegraphics[width=0.28\linewidth]{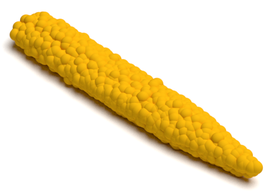} \\

Label: \textit{watercraft} & Label: \textit{airplane} & Label: \textit{airplane} \\
Pred: \textit{watercraft} & Pred: \textit{watercraft} & Pred: \textit{watercraft}
\end{tabular}
\caption{{\bfseries A target shape with wrong semantic interpretation.} Top row: a clean source point cloud, an adversarial example, and a clean target. Bottom row: corresponding reconstructions. Below the reconstructed point clouds, we indicate their shape class (Label) and the predicted label (Pred) by a classifier. Our attack may select a target shape that is classified wrongly.}
\label{fig:supp_classification}
\end{center}
\end{figure}

As noted in the paper, the attack may choose target point clouds that are misinterpreted semantically. Figure~\ref{fig:supp_confusion} presents the classification confusion matrix for the reconstructions of selected targets of the output space attack. In Figure~\ref{fig:supp_classification}, we show an example of a misclassification case.

The selection mechanism of a target point set is based on its geometric proximity to the source shape. Thus, the attack may select a non-standard instance from the target class. It leads to confusion modes between shape instances with similar characteristics, such as \textit{chair} and \textit{sofa} or \textit{airplane} and \textit{watercraft}, as raises from the confusion matrix in Figure~\ref{fig:supp_confusion}.


In Figure~\ref{fig:supp_classification}, a narrow \textit{airplane} instance is selected as a target. This is an unusual airplane shape, which has no wings. We note that for \textit{all} other visual attack examples in the paper and the supplementary material, the reconstructed point cloud is classified as the desired target shape category.

\subsection{Failure Cases} \label{sec:supp_failure_cases}
\begin{figure}[tb!]
\begin{center}
\includegraphics[width=\columnwidth]{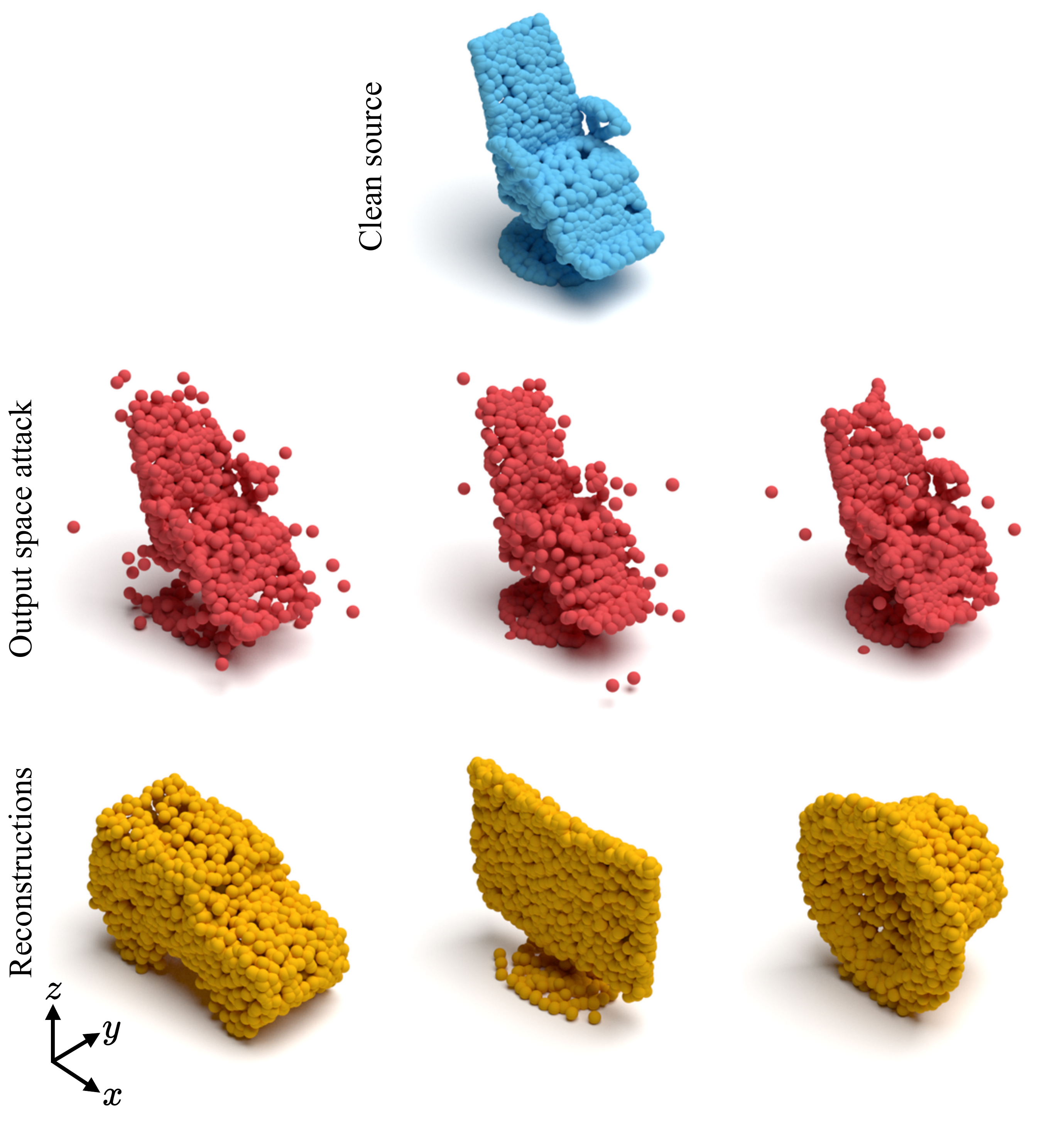}
\caption{{\bfseries Failure cases.} Top and middle rows: a clean \textit{chair} and its adversarial examples. Bottom row: corresponding reconstructions of the adversarial examples (a \textit{car}, a \textit{display}, and a \textit{loudspeaker}). The target shape is either too long on the $x$-axis, thin on the $y$-axis, or short on the $z$-axis, compared to the clean source. This geometric discrepancy results in distorted adversarial examples.}
\label{fig:supp_failure}
\end{center}
\end{figure}

Although our attack selects the nearest neighbors of the source from the target shape class, the geometry difference may still be large. In these cases, the attack causes high distortion to the source point cloud. Figure~\ref{fig:supp_failure} presents several failure examples for the output space attack.


\subsection{Reconstruction Error for Semantic Attack} \label{subsec:supp_recon_semantic_attack}
\begin{figure}[tb!]
\begin{center}
\begin{tabular}{ c c c }
Clean  & Semantic & Geometric     \\
source & attack~\cite{xiang2019generating}   & attack (ours) \\

\includegraphics[width=0.28\linewidth]{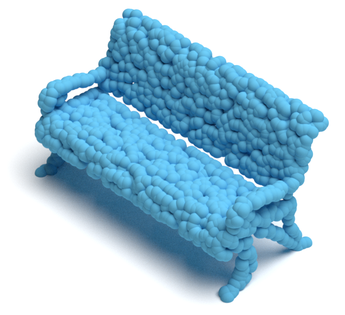} &
\includegraphics[width=0.28\linewidth]{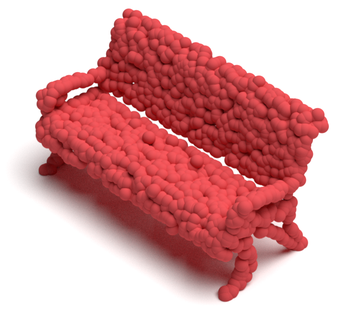} &
\includegraphics[width=0.28\linewidth]{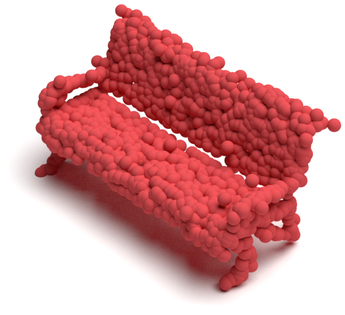} \\

\includegraphics[width=0.28\linewidth]{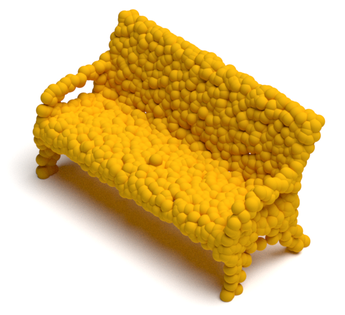} &
\includegraphics[width=0.28\linewidth]{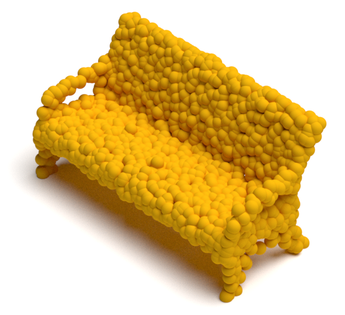} &
\includegraphics[width=0.28\linewidth]{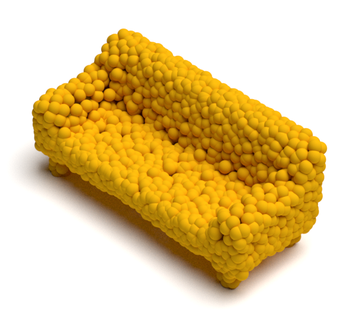} \\

$\SNRE = 1.00$ & $\SNRE = 1.08$    & $\SNRE = 6.13$    \\

\end{tabular}
\caption{
\textbf{Reconstruction comparison.} First row: a clean source point cloud (\textit{bench}), a semantic adversarial example (targeted to the \textit{sofa label}), and a geometric adversarial example of our output space attack (targeted to a \textit{sofa shape}). Bottom row: corresponding reconstructions and their source normalized reconstruction error, $\SNRE$. The reconstructed point cloud for the semantic attack is very similar to the reconstruction of the clean source shape, with an $\SNRE$ close to 1. In contrast, our attack changes the output geometry and results in a much higher $\SNRE$ value.}
\label{fig:supp_compare_snre}
\end{center}
\end{figure}

Our geometric attack aims to reconstruct a target \textit{point cloud instance}. Thus, we can evaluate the target normalized reconstruction error, $\TNRE$ (Equation~\ref{eq:t_nre}). In contrast, a semantic attack targets a classifier's prediction \textit{label}, and the $\TNRE$ measure is not applicable for such an attack. However, we can compute the source normalized reconstruction error, $\SNRE$ (Equation~\ref{eq:s_nre_before}). It measures the deviation from the reconstruction of the clean source point set by the AE. An $\SNRE$ of 1 indicates no deviation, while a high value implies a large difference. Figure~\ref{fig:supp_compare_snre} presents a visual example.

The semantic attack~\cite{xiang2019generating} that we used in the paper (sub-section~\ref{subsec:attack_evaluation}) resulted in an average $\SNRE$ of 1.32. In means that the reconstruction of its adversarial examples is very similar to that of the corresponding source point clouds. On the contrary, as reported in Table~\ref{tbl:defenses}, our attacks' $\SNRE$ is substantially higher. For example, it is 24.67 on average for the output space attack. We conclude that the semantic attack fails to alter the reconstructed geometry by the AE, while our attacks succeed.

\section{Additional Defense Results} \label{sec:supp_additional_defense}

\subsection{Defense for Untargeted Setting} \label{sec:supp_def_untargeted}
\begin{figure*}[tb!]
\begin{center}
\begin{tabular}{ c c c c c }
Clean  & Output space & Off-surface    & Critical points & Clean  \\
source & attack       & defense           & defense         & target \\

\includegraphics[width=0.17\linewidth]{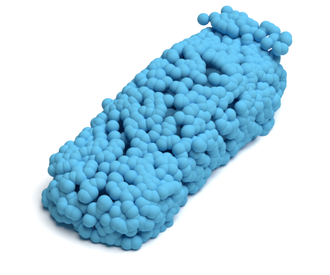} &
\includegraphics[width=0.17\linewidth]{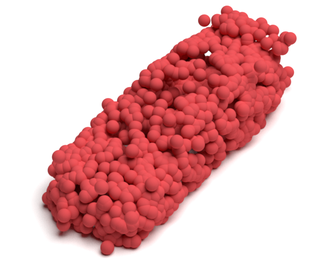} &
\includegraphics[width=0.17\linewidth]{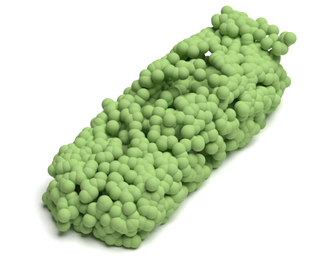} &
\includegraphics[width=0.17\linewidth]{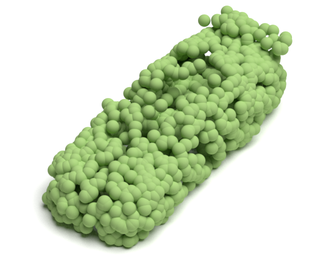} &
\includegraphics[width=0.17\linewidth]{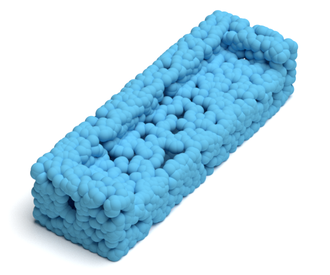} \\

\includegraphics[width=0.17\linewidth]{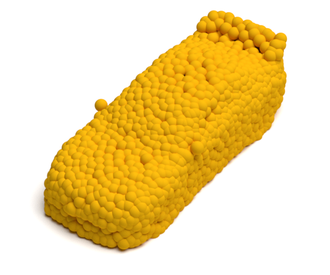} &
\includegraphics[width=0.17\linewidth]{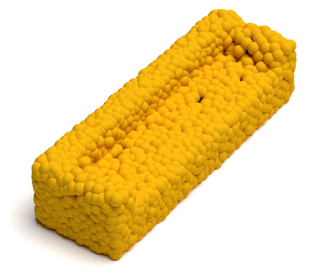} &
\includegraphics[width=0.17\linewidth]{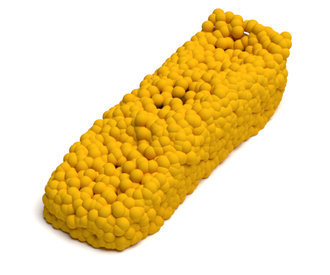} &
\includegraphics[width=0.17\linewidth]{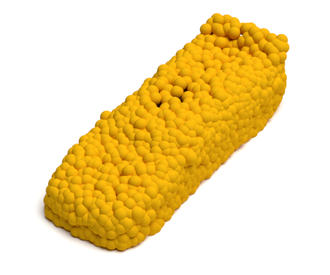} &
\includegraphics[width=0.17\linewidth]{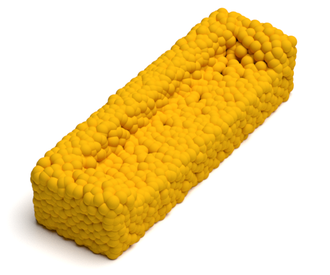} \\

$\SNRE = 1.00$ & $\SNRE = 4.11$    & $\SNRE = 1.54$ & $\SNRE = 1.51$ & N/A           \\
\end{tabular}
\caption{{\bfseries Defense for untargeted output space attack.} First row: a clean source point cloud (\textit{car}), an adversarial example of the untargeted output space attack, defended point cloud by the off-surface defense, defended point cloud by the critical points defense, and a clean target (\textit{sofa}). Second row: corresponding reconstructions and their source normalized reconstruction error, $\SNRE$ (not applicable for the reconstructed target). The defenses reduce the attack's influence and return the source shape's geometry at the output. However, a residual effect remains present in the resulting reconstructions.}
\label{fig:supp_defense_untargeted}
\end{center}
\end{figure*}

The paper reports defense results for the \textit{targeted} attack setting (Table~\ref{tbl:defenses}). Here, in Table~\ref{tbl:supp_defense_untargeted} and Figure~\ref{fig:supp_defense_untargeted}, we provide the defense's results for the \textit{untargeted} case. As explained in sub-section~\ref{subsec:supp_untargeted_attack_hist}, the untargeted attack causes a mild perturbation to the clean source and results in examples that are easier to defend against. In this case, after removing off-surface or critical points from the adversarial point cloud, the geometry and semantic interpretation of the source are obtained at the AE's output.

Albeit, there are still remnants of the attack in the reconstruction of the defended point set. As shown in Figure~\ref{fig:supp_defense_untargeted}, the output \textit{car} is flatter and narrower like the target \textit{sofa} shape. We conclude that our attack is not fully defendable, even in the untargeted setting.


\subsection{Decoder Side Defense} \label{subsec:supp_decoder_defense}
The defense is meant to do \textit{attack correction} on the \textit{encoder side}. Meaning, it removes adversarial points from the input to reduce the attack's effect on the output. However, a residual effect still exists, and features of the attack's target appear in the reconstructed shape after the defense. On the other hand, the defense has a negligible influence on a clean source point cloud, and its reconstruction after applying the defense remains practically the same.

This state gives rise to another layer of defense - \textit{attack detection} on the \textit{decoder side}. Its goal is to alert whether the reconstructed point cloud came from a tampered input. As a proof of concept, we did the following experiment. We split the source-target pairs of our attack into 76\%/8\%/16\% for train/validation/test sets, applied our defense on the clean source and adversarial examples, and ran them through the victim AE. Then, we trained a binary classifier on the reconstructed sets, where the labels for clean and adversarial point clouds were $0$ and $1$, respectively. The detection accuracy on the test set for our attacks and defenses is given in Table~\ref{tbl:supp_attack_detection}.

The baseline accuracy for the detection problem is 50\% (a random label choice). Our attack detection mechanism increases this accuracy for both attacks with both defenses. The detection accuracy is almost 70\% for the latent space attack, where for the output space attack it is about 60\%. We conclude that attack detection at the decoder side is possible to some extent. Nonetheless, it comes with the overhead of running the defense on the input and the detector on the output constantly.

\begin{table}[tb!]
\centering
\begin{tabular}{@{ } l c c c @{ }}
\toprule
Defense Type    &  $\SRE$ $\downarrow$     & $\SNRE$ $\downarrow$ & $\SRCA$ $\uparrow$ \\
\midrule
Surface (Lat)  & 4.06/1.26          & 9.04/2.49          & 28.6\%/77.5\%          \\
Critical (Lat) & 4.06/\textbf{1.16} & 9.04/\textbf{2.20} & 28.6\%/\textbf{81.9\%} \\
\midrule
Surface (Out)  &  4.11/0.94          & 8.88/1.76          & 31.4\%/85.5\%          \\
Critical (Out) &  4.11/\textbf{0.85} & 8.88/\textbf{1.52} & 31.4\%/\textbf{87.4\%} \\
\bottomrule
\end{tabular}
\vspace{0.2cm}
\caption{{\bfseries Geometric defense metrics and semantics for the untargeted setting before/after the defense.} The acronyms are the same as in Table~\ref{tbl:defenses}. $\SRE$ is multiplied by a factor of $10^3$. The arrows indicate whether a lower or higher value is better. The defenses are successful in reducing the source's reconstruction error and increasing its correct classification rate.}
\label{tbl:supp_defense_untargeted}
\end{table}

\section{Ablation Study} \label{sec:supp_ablation_study}

\newtxt{
\subsection{Random Target Selection} \label{subsec:supp_rand_target}
For a given source point cloud, we consider its geometric nearest neighbors from a target class as potential candidates for the attack. To examine the importance of this aspect, we selected target point clouds randomly for the output space attack. Its performance in this setting is provided in Table~\ref{tbl:supp_attacks_ablation}.

Random target selection increases the geometric discrepancy between the source and target point sets and compromises the attack's performance. As raises from the table, the number of $OS$ points is increased from $24$ to $36$, and the normalized reconstruction error is doubled. This result indicates the importance of the geometric target selection, as proposed in our attack.
}

\begin{table}[tb!]
\begin{center}
\begin{tabular}{ c | c c }
\diagbox{Attack}{Defense} & Surface & Critical \\
\hline
Latent & 68.0\% & 69.7\% \\
Output & 59.2\% & 61.0\% \\
\end{tabular}
\end{center}
\caption{{\bfseries Attack detection accuracy.} The accuracy is for detecting whether a reconstructed point cloud originates from a defended clean or a defended adversarial input. A higher value is better. Our attack detector increases the 50\% coin-flip accuracy by approximately 20\% and 10\% for the latent and output space attacks, respectively.}
\label{tbl:supp_attack_detection}
\end{table}

\subsection{Target Selection with Semantic Consideration} \label{subsec:supp_target_with_cls}
Our target selection mechanism is purely geometric. It chooses candidate point cloud instances from the target class according to their Chamfer Distance from the given source point cloud. Then, the point set that yields the lowest attack score (Equation~\ref{eq:attack_score}) is selected for the corresponding source. As reported in Table~\ref{tbl:attack_semantics} in the paper, this selection method results in a relatively low classification accuracy for reconstructed targets and adversarial examples.

As an alternative, we restricted the selection to a subset of targets that a classifier recognizes correctly. In turn, it substantially improved the semantic performance of the attack. We highlight here that this change influences the attack in other aspects. First, the classifier becomes a part of the method rather than a means to its evaluation. Now, access to the classifier, which was trained on the same shape categories used for the AE's training, is required.


Second, it affects the attack's geometric performance since the classifier rules out potential instances from the target class. Table~\ref{tbl:supp_attacks_ablation} presents the results for this case. Compared to Table~\ref{tbl:attacks_loss_dist_chamfer}, the normalized error is higher by $3\%$ ($1.14$ instead of $1.11$). For the latent space attack, this setting resulted in a $\TNRE$ of $2.38$ instead of $2.16$ (a 22\% increase).

To conclude, in the paper, we preferred a clear separation between geometric and semantic considerations when performing the attack. However, if a classifier is assumed to be available, it can be beneficial, in the semantic aspect, to leverage its predictions when selecting target point clouds for the attack at the expense of some geometric performance reduction.


\begin{table}[tb!]
\centering
\begin{tabular}{@{ } l c c c c @{ }}
\toprule
Attack Type  & \#$OS$  & $\SCD$ & $\TRE$ & $\TNRE$  \\
\midrule
Output (A) & 36  & 0.75 & 0.84 & 2.01 \\
Output (B) & 26  & 0.48 & 0.66 & 1.14 \\
Output (C) & 28  & 0.47 & 0.80 & \textbf{1.06} \\
Output (D) & 28  & \textbf{0.39} & 0.78 & 1.38 \\
Output (E) & 30  & 0.56 & 0.91 & 1.58 \\
\midrule
Proposed   & \textbf{24} & 0.44 & \textbf{0.63} & 1.11 \\
\bottomrule
\end{tabular}
\vspace{0.2cm}
\caption{{\bfseries Ablation study for the output space attack.} Geometric adversarial metrics are reported for different variations of our attack (one change at a time): (A) selecting targets at random, (B) a selection of targets with a correct semantic prediction , (C) one target candidate per source, (D) direct perturbation restriction, and (E) the addition of an off-surface loss. The acronyms are the same as in Table~\ref{tbl:attacks_loss_dist_chamfer} in the paper. $\SCD$ and $\TRE$ are multiplied by a factor of $10^3$. A lower value is better. Compared to the proposed method, the ablative settings either increase the source distortion, compromise the reconstruction quality, or both. Please see Section~\ref{sec:supp_ablation_study} for more details.}
\label{tbl:supp_attacks_ablation}
\end{table}


\subsection{Number of Target Candidates} \label{subsec:supp_candidates_number}
For each source point cloud in our attack, we have $5$ candidate instances from the target shape class. As an ablation test, we ran our attack with one target candidate only (the first nearest neighbor). As seen in Table~\ref{tbl:supp_attacks_ablation}, this experiment increased both the source distortion $\SCD$ and target reconstruction error $\TRE$, compared to the results in the paper (Table~\ref{tbl:attacks_loss_dist_chamfer}).

We note that in this case, the relative error ($\TNRE$) is $1.06$, smaller than that reported in the paper, which was $1.11$. However, the absolute error ($\TRE$) is much higher, and thus, this setting is undesired. To summarize, using several potential targets per source enables the attack to achieve better geometric performance.



\subsection{Direct Perturbation Restriction} \label{subsec:supp_perturb_restrict}
In the optimization process of our attack, we employed Chamfer Distance to preserve geometric proximity between the clean source and the adversarial point sets. An alternative approach is to restrict the perturbation norm directly by an $L_2$ loss. In this case, Equation~\ref{eq:loss_dist} takes the form:
\begin{equation} \label{eq:supp_loss_pert}
\Lagr_{distance}(Q, S) = ||Q - S||_2 = ||P||_2.
\end{equation}

We optimized our attack with the perturbation loss and recalibrated $\lambda$ to level it with the adversarial loss in Equation~\ref{eq:problem_def_output}. The results of this experiment are provided in Table~\ref{tbl:supp_attacks_ablation}. Compared to the adversarial metrics in Table~\ref{tbl:attacks_loss_dist_chamfer}, almost all the metrics are compromised when the perturbation loss is employed. Similarly, using this loss for the latent space attack raised the number of $OS$ points to $38$ and the $\TNRE$ to $2.53$.

The perturbation loss penalizes perturbed points according to their distance from the original corresponding point in the clean set and does not consider the shape's geometry. Specifically, it does not account for proximity to other points in the set, as may occur for a shift tangential to the surface. We conclude that direct perturbation restriction hinders the attack's search space, and therefore find it less suitable for our purposes.

\subsection{Off-Surface Loss} \label{subsec:supp_surface_loss}
Our distance loss (Equation~\ref{eq:loss_dist}) averages the distance between each point in the adversarial example and its nearest neighbor point in the source point set. Thus, the attack may craft an example with mostly on-surface points and several off-surface ones that do not have a close neighbor. In an attempt to minimize the number of off-surface ($OS$) points during the optimization process of the attack, we added the maximal nearest neighbor distance to its optimization objective. In this experiment, Equation~\ref{eq:loss_dist} was changed to:
\begin{equation} \label{eq:supp_loss_surf}
\begin{split}
\Lagr_{distance}(&Q, S) = \\
&CD(Q, S) + \beta \max_{q \in Q}{\min_{s \in S}||q - s||_2^2},
\end{split}
\end{equation}

\noindent where $\beta$ is a hyperparameter.

A low value of $\beta = 0.001$ did not influence the number of $OS$ points. However, a high $\beta$ value did not reduce their number but rather increased it and also compromised the reconstruction quality of the attack's target. For example, setting $\beta = 0.1$ resulted in $30$ $OS$ points and a $\TNRE = 1.58$ for the output space attack (see Table~\ref{tbl:supp_attacks_ablation}). Similar to direct perturbation restriction (sub-section~\ref{subsec:supp_perturb_restrict}), the maximal nearest neighbor loss limits the quest of the attack for adversarial solutions with good geometric performance, and thus, we did not use it in our method. This experiment's result suggests that our attack, as presented in the paper, crafts adversarial examples with both a low number of off-surface points and a low reconstruction error of the target point cloud.




\subsection{Latent Nearest Neighbors} \label{subsec:supp_latent_nn}
In the paper, we selected candidates for attack according to the distance in the point space for both the latent and output space attacks (as explained at the beginning of Section~\ref{sec:method}). A reasonable alternative for the latent space attack is to choose candidates according to the distance in the latent space. To examine this option, we ran the attack, where the target for each source was selected from the Euclidean nearest neighbors in the latent space.

The adversarial metric results for this experiment were the same as those reported in Table~\ref{tbl:attacks_loss_dist_chamfer} in the paper. It happened since the proximity \textit{ranking} of geometric nearest neighbors is mostly preserved in the AE's latent space. Thus, selecting neighbors in the latent space is equivalent to the selection in the point space in terms of the attack's performance. We decided to use geometric neighbors for both the latent and output space attacks to enable a direct comparison between them.

\subsection{Off-Surface Defense Calibration} \label{subsec:supp_oos_calib}
The off-surface defense requires the calibration of the number of nearest neighbors $k$ for computing the average distance per point and the distance threshold $\delta$. A small $k$ value will result in a noisy distance measure. If $k$ is too large, the average distance will be less discriminative for detecting off-surface points. Regarding the threshold, setting a high value may not filter all the adversarial points, while a small value will filter on-surface points and compromise the reconstruction quality.

We applied the defense for the output space attack with different $k$ and $\delta$ values to obtain a minimal reconstruction error for the defended input, with respect to the clean source point cloud ($\SNRE_{after}$, Equation~\ref{eq:s_nre_after} in the paper). The results appear in Table~\ref{tbl:supp_surface_def_calib}. We found that $k=2$ gives a discriminative distance measure and $\delta=0.04$ balances between keeping on-surface points while filtering out-of-surface ones. Thus, this calibration results in the lowest $\SNRE_{after}$. We note that this calibration applies to the latent space attack as well.

\begin{table}[tb!]
\begin{center}
\begin{tabular}{ c | c c c c }
\diagbox{$k$}{$\delta$} & $0.03$ & $0.04$ & $0.05$ & $0.06$ \\
\hline
1 & 3.16   & 3.11          & 3.29 & 3.70 \\
2 & 3.34   & \textbf{3.10} & 3.19 & 3.47 \\
4 & 3.70   & 3.27          & 3.20 & 3.37 \\
8 & 4.28   & 3.61          & 3.35 & 3.38 \\
\end{tabular}
\end{center}
\caption{{\bfseries Off-surface defense calibration.} The defense is applied with different parameter settings to the output space attack. The $\SNRE_{after}$ (Equation~\ref{eq:s_nre_after}) is computed for each setting. A lower value is better. Selecting the number of neighbors $k=2$ and the distance threshold $\delta=0.04$ gives the lowest error.}
\label{tbl:supp_surface_def_calib}
\end{table}



\begin{figure}[tb!]
\begin{center}
\begin{tabular}{ c c c }
\whitetxt{aa}\textit{Bookshelf} & \whitetxt{aa}Attack & \whitetxt{aaa}\textit{Bottle} \\
\includegraphics[width=0.20\linewidth]{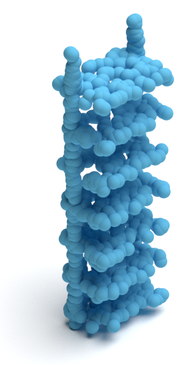} &
\includegraphics[width=0.20\linewidth]{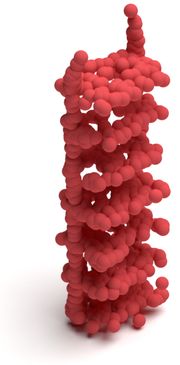} &
\includegraphics[width=0.20\linewidth]{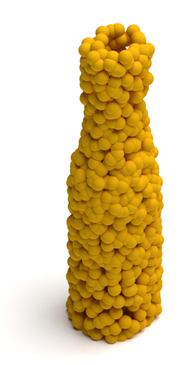} \\

\\
\whitetxt{aaa}\textit{Bottle} & \whitetxt{aaa}Attack & \whitetxt{aaa}\textit{Vase} \\
\includegraphics[width=0.26\linewidth]{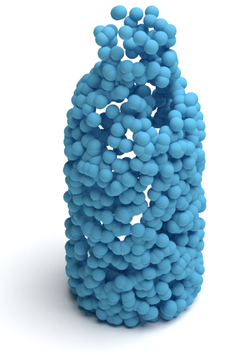} &
\includegraphics[width=0.26\linewidth]{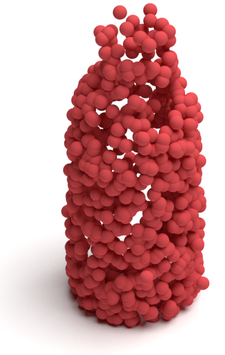} &
\includegraphics[width=0.26\linewidth]{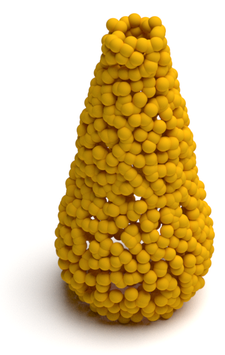} \\

\\
\textit{Chair} & Attack & \textit{Toilet} \\
\includegraphics[width=0.28\linewidth]{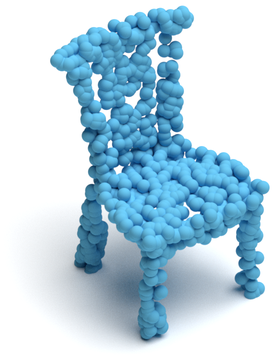} &
\includegraphics[width=0.28\linewidth]{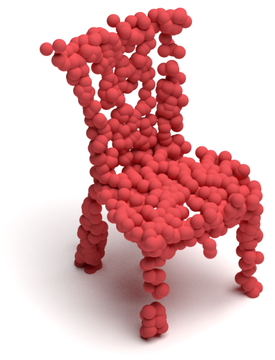} &
\includegraphics[width=0.28\linewidth]{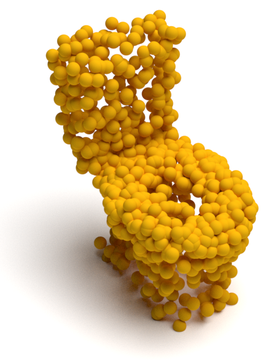} \\

\\
\whitetxt{aaaa}\textit{Vase} & \whitetxt{aaa}Attack & \whitetxt{aaa}\textit{Bottle} \\
\includegraphics[width=0.28\linewidth]{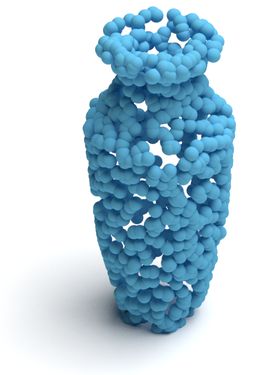} &
\includegraphics[width=0.28\linewidth]{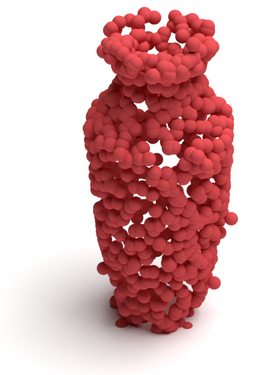} &
\includegraphics[width=0.28\linewidth]{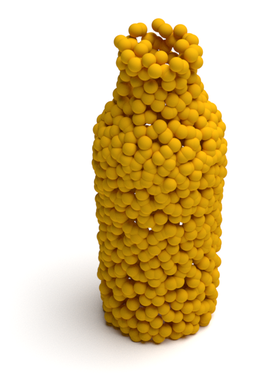} \\
\end{tabular}
\caption{\textbf{Attack results for ModelNet40 dataset.} In each row, we show from left to right a clean source point cloud, an adversarial example of the output space attack, and its reconstruction by the victim AE. Above the source and reconstructed point clouds, we indicate their shape class. Our attack successfully transforms a shape from one class into a shape from a different class.}
\label{fig:supp_attack_modelnet}
\end{center}
\end{figure}

\section{Attack for ModelNet40 Dataset} \label{sec:supp_attack_modelnet}
Our attack framework is not limited to a specific dataset nor specific shape classes. To demonstrate this property, we apply our attack to the ModelNet40 dataset~\cite{wu2015modelnet}. We use point clouds with 1024 points~\cite{xiang2019generating} uniformly sampled from the dataset's models and normalize them to the unit cube. Then, we train a victim AE network on the official train-split data. The AE has the same architecture as the one used in the paper for the ShapeNet Core55 dataset~\cite{chang2015shapenet}.


Since ModelNet40 is a highly imbalanced dataset, we use its ten largest categories and randomly select 25 instances per category, unseen by the AE during training. Then, we run the output space attack and measure its geometric performance. Visual examples of the attack are presented in Figure~\ref{fig:supp_attack_modelnet}.

Similar to the results reported in Table~\ref{tbl:attacks_loss_dist_chamfer} in the main body, the attack resulted in 21 off-surface ($OS$) points and a $\TNRE$ of 1.13. It indicates the high performance of the attack for ModelNet40 data. We note that this experiment included shape categories other than those used in the paper, such as \textit{bookshelf}, \textit{bottle}, \textit{vase}, and \textit{toilet}. This experiment suggests that our attack method is flexible and can be applied to different datasets with various shape classes.

\section{Experimental Settings} \label{sec:supp_experimental_sett}

\subsection{Victim Autoencoder} \label{subsec:sup_victim_ae}
We adopt the published AE architecture of Achlioptas \etal~\cite{achlioptas2018learning}. The encoder has $5$ per-point convolution layers, with filter sizes $(64, 128, 128, 256, m)$. Each layer includes batch normalization and ReLU operations. The last layer is followed by a global feature-wise max-pooling operation to obtain a latent feature-vector of size $m = 128$. The decoder is a series of $3$ fully-connected layers, with sizes $(256, 256, n \times 3)$, where $n$ is the number of output points (either 2048 for ShapeNet Core 55 dataset or 1024 for ModelNet40 dataset). The first two layers of the decoder include ReLU non-linearity. We train the AE with Chamfer Distance loss (Equation~\ref{eq:chamfer_dist} in the paper) and Adam optimizer with momentum $0.9$, as recommended by the authors. Additional settings appear in Table~\ref{tbl:supp_ae_training_settings}.

\subsection{Attack Optimization} \label{aubsec:supp_attack_optim}
Table~\ref{tbl:supp_attack_optim_params} summarizes the optimization parameters for our attacks. We ran the optimization for $500$ gradient steps. Starting from iteration $400$, near the convergence point of the attack, we checked the target reconstruction error and kept the result with the minimal error. On average, the optimization time for a single input point cloud was $0.69$ seconds for the latent space attack and $0.72$ seconds for the output space attack, when running the attacks on an Nvidia Titan Xp GPU.

\subsection{Autoencoders for Attack Transfer} \label{subsec:supp_aes_for_transfer}
We trained an AE with the same architecture as the attacked one, as detailed in sub-section~\ref{subsec:sup_victim_ae}. The training settings were also the same, up to a different random weight initialization. For the other AEs for the transfer experiment, we followed the official implementation~\cite{groueix2018atlasnet} or published paper~\cite{yang2018folding} of the authors.

AtlasNet~\cite{groueix2018atlasnet} employs a PointNet-based encoder. It has a multi-layer perceptron (MLP) of size $(64, 128, 1024)$, followed by a global max-pooling over points and fully-connected layers of sizes $(1024, 1024)$. Each encoder layer has batch normalization and ReLU activation, except the last MLP layer that is without batch normalization. The encoder produces a global latent vector of size $1024$.

The decoder operates on sets of 2D points from the unit square. First, it applies a per-point filter of size $1024$ and adds the global latent vector to each point. Then, it uses an MLP of size $(512, 512, 512, 3)$ to obtain the output 3D point cloud. All the layers include batch normalization and ReLU, except for the last layer. The decoder runs on $25$ 2D unit squares, with $100$ points at each square, and outputs $2500$ points in total.

FoldingNet~\cite{yang2018folding} takes as input $xyz$ coordinates and the local covariance for each point. Its encoder contains an MLP of $3$ layers, with filter sizes $(64, 64, 64)$. Then, two graph layers aggregate information from $16$ neighboring points and increase the number of point features to $1024$, with an MLP of size $(128, 1024)$. A global max-pooling and additional $2$ fully-connected layers of sizes $(512, 512)$ produce a $512$-dimensional latent code-word. Each layer of the encoder, except the last one, includes batch normalization and ReLU activation.

\begin{table}[tb!]
\centering
\begin{tabular}{@{ } l c c c c @{ }}
\toprule
&  MLP~\cite{achlioptas2018learning} & AtlasNet~\cite{groueix2018atlasnet} & FoldingNet~\cite{yang2018folding} \\
\midrule
TEs & $500$    & $150$   & $25$     \\
BS  & $50$     & $32$    & $8$      \\
LR  & $0.0005$ & $0.001$ & $0.0001$ \\
TT  & $6$      & $7$     & $153$    \\
\bottomrule
\end{tabular}
\vspace{0.2cm}
\caption{{\bfseries Training settings of the victim AE and the AEs for attack transfer.} MLP refers to the attacked AE. AtlasNet and FoldingNet are the AEs for transfer. TEs, BS, LR, and TT stand for training epochs, batch size, learning rate, and training time, respectively. The training time is reported in hours for an Nvidia Titan Xp GPU.}
\label{tbl:supp_ae_training_settings}
\end{table}
\begin{table}[tb!]
\centering
\begin{tabular}{@{ } l c @{ }}
\toprule
Parameter &  Value \\   
\midrule
Gradient steps                  & $500$ \\
Learning rate                   & $0.01$ \\
$\lambda$ Perturbation/Latent   & $0.15$ \\
$\lambda$ Perturbation/Output   & $0.002$ \\
$\lambda$ Chamfer/Latent        & $150$ \\
$\lambda$ Chamfer/Output        & $1$ \\
\bottomrule
\end{tabular}
\vspace{0.2cm}
\caption{{\bfseries Optimization parameters.} The table details parameter values for our attacks. We indicate the $\lambda$ parameter for different distance/adversarial loss type settings. Perturbation and Chamfer refer to Equations~\ref{eq:supp_loss_pert} and \ref{eq:loss_dist}, respectively. Latent and Output refer to Equations~\ref{eq:loss_latent} and \ref{eq:loss_output}, respectively.}
\label{tbl:supp_attack_optim_params}
\end{table}

The decoder concatenates the code-word to 2D grid points and folds them two times to reconstruct the 3D shape. Each fold is implemented by an MLP of size $(512, 512, 3)$, with ReLU non-linearity in the first two layers of the MLP. The 2D grid contains $2025$ points, which is the number of 3D output points.

We trained the AEs with Chamfer Distance loss (Equation~\ref{eq:chamfer_dist}), as employed for the victim AE. Additional training settings are given in Table~\ref{tbl:supp_ae_training_settings}.

\subsection{Classifier for Semantic Evaluation} \label{subsec:supp_classifier_semantic}
We used PointNet architecture with $13$ output classes, the number of shape classes for our attack. We trained the network with the same settings as Qi \etal~\cite{qi2017pointnet}, with $150$ epochs instead of $250$ and without random point cloud rotation during training. These modifications improved the classification performance in our case. Training the classifier took about $18$ hours.

\subsection{Classifier for Decoder Side Defense} \label{subsec:supp_classifier_decoder_defense}
The classifier for this experiment (sub-section~\ref{subsec:supp_decoder_defense}) was a PointNet with $2$ output labels. We used the same training settings as Qi \etal~\cite{qi2017pointnet}, except for a random rotation during training and a different number of epochs. For the latent and output space attacks, with both defenses, we trained the classifier for $80$ and $100$ epochs, respectively. The training took approximately $1$ and $1.25$ hours for the corresponding classifier instance.

\end{document}